\documentclass[twoside]{article}
\usepackage{graphicx} 
\usepackage{xspace}

\usepackage{amsmath,amsfonts,amscd,amssymb,amsthm}
\usepackage[scr=rsfs]{mathalpha}
\usepackage{graphicx}
\usepackage{epstopdf}
\usepackage{mathtools}
\usepackage{bm}
\usepackage{microtype}

\usepackage{booktabs}
\usepackage{array}
\usepackage{multirow}
\usepackage{caption}
\usepackage{subcaption}
\usepackage[table]{xcolor}

\usepackage{wrapfig}
\usepackage{lipsum}

\usepackage[colorlinks = true,
            linkcolor = blue,
            urlcolor = blue,
            citecolor = blue,
            anchorcolor = blue]{hyperref}       
\usepackage{url}
\usepackage{fancyhdr}
\usepackage{multirow}

\usepackage{enumitem}

\usepackage{float}

\usepackage{subeqnarray}

\usepackage{algorithm}
\usepackage{algorithmic}

\usepackage{standalone}
\usepackage{tikz}
\usetikzlibrary{positioning,calc}
\usepackage{pgfplots}
\pgfplotsset{compat=1.17}
\usepgfplotslibrary{groupplots,fillbetween}
\usepackage{xspace}

\newtheorem{remark}{\bf Remark}[section]

\definecolor{rs_blue}{cmyk}{1,.10,0,.10}

\definecolor{lqr_col}{RGB}{255, 190, 11}
\definecolor{skvi_col}{RGB}{251, 86, 7}
\definecolor{sac_q_col}{RGB}{255, 0, 110}
\definecolor{sac_v_col}{RGB}{131, 56, 236}
\definecolor{sakc_col}{RGB}{58, 134, 255}

\newcommand{\greyrule}{\arrayrulecolor{black!20}\midrule\arrayrulecolor{black}}

\newcommand{\ack}[1]{\section*{Acknowledgements}#1}

\usepackage[margin={2cm,2cm}]{geometry}
\usepackage{scalerel}

\providecommand{\keywords}[1]
{
  \small	
  \textbf{\textit{Keywords---}} #1
}

\title{\LARGE\textbf{
    Koopman-Assisted Reinforcement Learning}
}
\author{
  {\bf
    Preston Rozwood$^{{\color{blue}\boldsymbol{\alpha}}, \dagger, *}$, %
    Edward J. Mehrez$^{{\color{blue}\boldsymbol{\beta}}, {\color{blue}\boldsymbol{\gamma}}, \dagger, *}$, %
    Ludger Paehler$^{{\color{blue}\boldsymbol{\delta}}}$, %
  } \\
  {\bf
    Wen Sun$^{{\color{blue}\boldsymbol{\alpha}}}$, %
    Steven L. Brunton$^{{\color{blue}\boldsymbol{\zeta}}, *}$, %
  } \\
  {\normalsize{
    $^{\alpha}$~Department of Computer Science, Cornell University, Ithaca, NY 14850, USA}
  }\\
  {\normalsize{
    $^{\beta}$~Subharmonic Technologies Inc., Fort Lauderdale, FL 33304, USA}
  }\\
  {\normalsize{
    $^{\gamma}$~Department of Economics, Cornell University, Ithaca, NY 14850, USA}
  }\\
  {\normalsize{
    $^{\delta}$~Department of Engineering Physics and Computation, Technical University of Munich,}
  }\\
  {\normalsize{
    Garching b. Muenchen, 85748, GER}
  }\\
  {\normalsize{
    $^{\zeta}$~Department of Mechanical Engineering, University of Washington, Seattle, WA 98195, USA}
  }
}
\date{}

\begin{document}

\maketitle
\def\thefootnote{$\dagger$}\footnotetext{Equivalent Contributions.}\def\thefootnote{\arabic{footnote}}
\def\thefootnote{*}\footnotetext{Corresponding Authors: Preston Rozwood(\href{mailto:pwr@cornell.edu}{pwr@cornell.edu}), Edward Mehrez (\href{mailto:ejm322@cornell.edu}{ejm322@cornell.edu}) and Steven L. Brunton (\href{mailto:sbrunton@uw.edu}{sbrunton@uw.edu})}\def\thefootnote{\arabic{footnote}}

\begin{abstract}
    The Bellman equation and its continuous form, the Hamilton-Jacobi-Bellman equation, are ubiquitous in reinforcement learning and control theory.
    However, these equations become intractable for high-dimensional or nonlinear systems. 
    This paper develops two new reinforcement learning algorithms based on the data-driven Koopman operator, which lifts a nonlinear system into new coordinates where the dynamics become approximately linear, and where Hamilton-Jacobi-Bellman-based methods are more tractable.
    In particular, the Koopman operator captures the expectation of the time evolution of the value function via linear dynamics in the lifted coordinates.
    By parameterizing the Koopman operator with the control actions, we construct a ``controlled Koopman tensor'' that facilitates the estimation of the optimal value function.
    This enables us to reformulate two max-entropy RL algorithms: soft value iteration and soft actor-critic.
    This flexible and interpretable framework includes deterministic and stochastic systems, as well as discrete and continuous dynamics.

    Koopman Assisted reinforcement learning attains state-of-the-art performance with respect to traditional neural network-based soft actor-critic baselines on a linear state-space system, the Lorenz system, fluid flow past a cylinder, and a double-well potential with non-isotropic stochastic forcing.
\end{abstract}


%
%
%
%

\section{Introduction}

Reinforcement learning (RL) is a rapidly developing field at the intersection of machine learning and control theory, in which an intelligent agent learns how to interact with a complex environment to achieve an objective~\cite{Sutton1998book,Brunton2022data}.  
Deep reinforcement learning (DRL)~\cite{lillicrap2015continuous,mnih2016asynchronous,van2016deep,wang2016dueling,ravichandiran2018hands,hessel2018rainbow,reddy2018shared} has demonstrated human-level or super-human performance in several challenging tasks, including video games~\cite{mnih2015human,vinyals2019grandmaster} and strategy games~\cite{silver2016mastering,silver2017mastering,silver2018general}.  DRL is also increasingly used for scientific and engineering applications, including for drug discovery~\cite{popova2018deep}, robotic manipulation~\cite{gu2017deep}, autonomous driving~\cite{sallab2017deep}, drone racing~\cite{kaufmann2023champion}, fluid flow control~\cite{gazzola2014reinforcement,colabrese2017flow,verma2018efficient,novati2019controlled,biferale2019zermelo,fan2020reinforcement,bae2022scientific,lagemann2025hydrogyma,lagemann2025hydrogymb}, and fusion control~\cite{degrave2022magnetic}.  
Despite this progress, DRL solutions often require tremendous computational resources to train and generalize, and typically lack the interpretability critical in many applications.
The algorithms underlying many reinforcement learning strategies are closely related to the Bellman and Hamilton-Jacobi-Bellman equations from optimization and optimal control theory~\cite{Brunton2022data}.  However, solving these equations becomes intractable for high-dimensional, nonlinear systems, which often motivates the use of deep learning and other surrogate modeling approaches~\cite{retchin2023koopman,zolman2024sindy}.   
In this paper, we re-examine the Bellman equation through a novel application of the Koopman operator~\cite{koopman1931hamiltonian,koopman1932dynamical,Mezic2004physicad,mezic2005spectral,Budivsic2012chaos,Mezic2013arfm,Brunton2022siamreview}, which allows us to recast a nonlinear dynamical system as a linear system of equations on an infinite-dimensional function space. Given a deterministic, discrete-time dynamical system
\begin{align}
    x' = {F}(x)
\end{align}
the Koopman operator provides an alternative perspective in which these dynamics become linear on an infinite-dimensional Hilbert space of measurement functions of the system. Mathematically, given a function from the state space to the reals $g:\mathcal{X}\to \mathbb{R}$, the Koopman operator $\mathcal{K}$ is defined for deterministic (time-homogenous) autonomous systems as:
\begin{equation}
    \mathcal{K}g(x) := g({F}(x)) =  g(x').
\end{equation}
More generally, in stochastic autonomous systems, it is defined as the conditional forecast operator:
\begin{align}
    \mathcal{K} g(x) &= \mathbb{E}\left(g(X')|X=x\right).
\end{align}

\noindent For a system with control $u$, the dynamics may be written as
\begin{equation}
    x' = F(x,u)
\end{equation}
and a given reinforcement learning policy $\pi(u|x)$ determines the probability of taking a control action $u$ given a state $x$.  
For a deterministic system, the value function for a policy $\pi$ is given by:
\begin{align}
    V^{\pi}(x) = r(x, u) + \gamma V^\pi (x'),
\end{align}
which is particularly challenging because it involves the future state $x'$ which is nonlinearly related to $x$ and $u$ via $F$. This paper recasts the continuation term in the Bellman equation in terms of the Koopman operator, specifically viewing the value function $V^\pi(x)$ for a given policy $\pi$ as the Koopman observable.  The value function may be rewritten using the Koopman operator as:
\begin{align}
    V^{\pi}(x) = r(x, u) + \gamma \mathcal{K} V^\pi (x). 
\end{align}
In this Koopman formulation, the value function only relies on the current state $x$, whereas in the original formulation it depends on the future state $x'$; here, $r$ is the reward function and $\gamma$ is a discount factor. For a stochastic system the value function transforms according to

\begin{align}
    &V^{\pi}(x) = \mathbb{E}_{u\sim\pi(\cdot|x)}\left\{r(x, u) + \gamma \mathbb{E}_{x' \sim p(\cdot|x,u)} \left[V^\pi (x')\right]\right\} \notag\\
    \Longrightarrow \quad
    &V^{\pi}(x) = \mathbb{E}_{u\sim\pi(\cdot|x)}\left\{r(x, u) + \gamma \mathcal{K}V^\pi(x)\right\}. \label{basicKoopmanBellman}
\end{align}

\begin{figure}[t]
    \centering
    \includegraphics[width=\columnwidth]{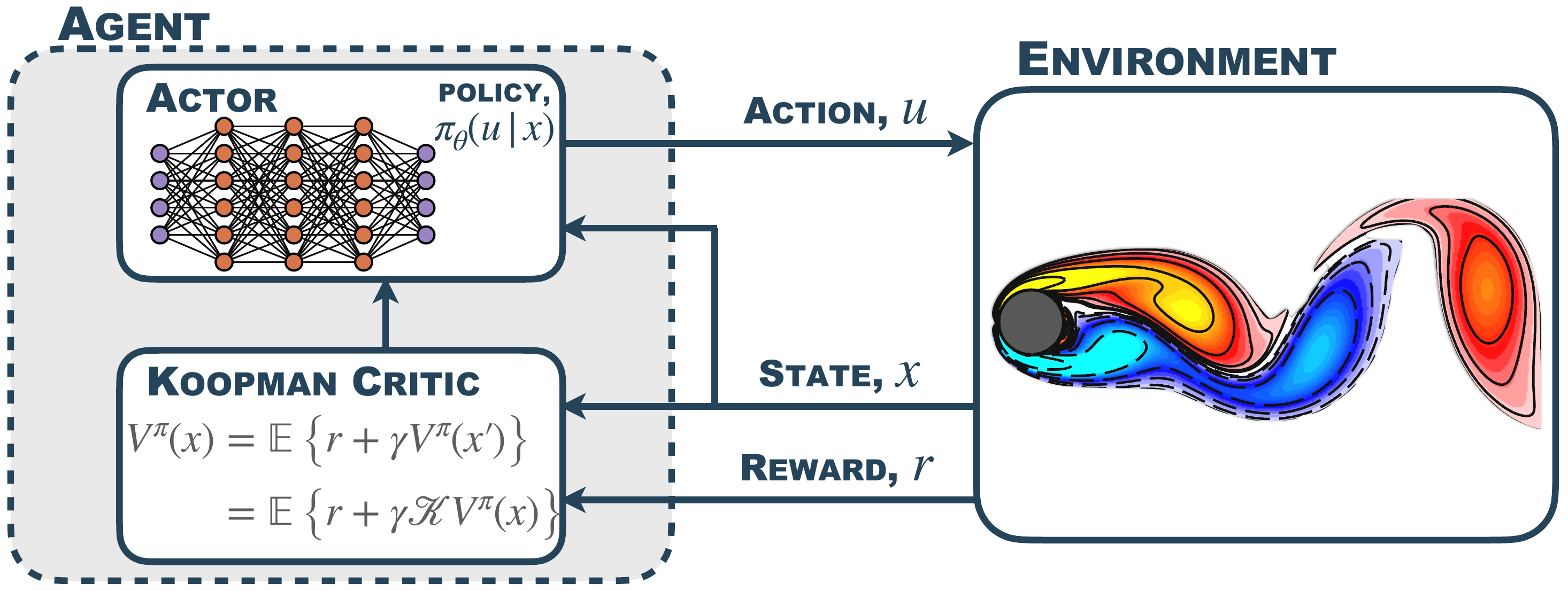}
    \caption{
        Koopman-assisted reinforcement learning in the example of the Soft Actor Koopman-Critic, a Koopman variant of the popular Soft Actor-Critic algorithm. The \textit{Koopman Critic} receives the state and the reward in the original state-space, before lifting these variables to a feature space, where the value function can be advanced in time with the \textit{Koopman operator}. This critique is then fed back to the Actor which issues the action to be performed.  This is 1 of 2 main Koopman-assisted reinforcement learning algorithms explored in this work; the other algorithm is a modified soft Koopman value iteration.
    }
    \label{fig:soft-actor-koopman-critic}
\end{figure}

\noindent When considering a particular observable, such as the value function in reinforcement learning, it may be possible to learn a finite-dimensional basis, or dictionary, in which this function and its evolution via the Koopman operator can be well approximated. There are many approaches to approximate the Koopman operator from data, such as dynamic mode decomposition (DMD)~\cite{Rowley2009jfm,schmid2010dynamic,Tu2014jcd,Kutz2016book,Askham2018siads,azencot2019consistent}, the extended dynamic mode decomposition (EDMD)~\cite{Williams2015jnls,Williams2015jcd,folkestad2020extended,colbrook2023mpedmd}, and related robust extensions~\cite{colbrook2023beyond,colbrook2023residual,colbrook2024rigorous}. This body of work was subsequently extended to stochastic dynamical systems~\cite{klus2018data,klus2020data} and controlled dynamical systems~\cite{proctor2016dynamic,proctor2018generalizing,kaiser2021data}. The correct choice of dictionary space is crucial in Koopman analysis techniques since they are generally not robust to misspecification.

\subsection{Related Work}
\label{sec:related_work}

The Koopman operator has recently been applied to RL, with several noteworthy developments. 
Several works have used the Koopman operator for advancing $Q$ learning, including developing a modified deep $Q$ network (DQN)~\cite{sinha2022koopman} using imitation learning, and Koopman $Q$-learning based on identifying symmetries in state dynamics~\cite{weissenbacher2022koopman}. 
Retchin et al.~\cite{retchin2023koopman} developed a Koopman constrained policy optimization (KCPO), combining a Koopman autoencoder network with differentiable model predictive control (MPC). 
Although Koopman autoencoders have been explored for dynamical systems for several years~\cite{lusch2018deep,morton2018deep,otto2019linearly,Takeishi2017nips,yeung2019learning,Mardt2018natcomm}, their application in control is recent.
More generally, several studies have extended Koopman analysis from autonomous systems to systems with actuation and control~\cite{proctor2016dynamic,proctor2018generalizing,kaiser2021data}.  However, these studies did not extend the Koopman embedding framework to the setting of Markov decision processes (MDPs), which is a main contribution of this present work.  
Notably, we introduce a new approach to parameterize the Koopman operator using a Koopman tensor on a lifted state-control space, which is essential for handling controlled dynamical systems. The Koopman tensor approach builds upon and extends methods like DMDc~\cite{proctor2016dynamic} and Koopman with inputs and control (KIC)~\cite{proctor2018generalizing} by capturing a wider range of interactions between states and actions. DMDc restricts these interactions to be purely linear and additive, limiting its ability to model complex dynamics; however, the Koopman tensor utilizes Kronecker products between any dictionary spaces, enabling it to capture rich multiplicative and nonlinear relationships. This enhanced flexibility leads to improved model accuracy and allows the Koopman tensor to effectively handle systems with intricate control dependencies.  KIC is more general than DMDc, although it requires the prediction of future actions or forcing their predicted values to be zero, which can lead to inaccuracies in predictions of other values. 
The related sparse identification of nonlinear dynamics with control (SINDYc) algorithm~\cite{brunton2016sparse,Kaiser2018prsa} also models controlled dynamical systems. SINDYc offers high flexibility through its generality on the candidate function space for states and controls. In particular, feature maps richer than the multiplicative separable case we consider could be considered with SINDYc.  

Some of the earliest work leveraging data-driven Koopman models to design controllers involved the use of DMD and eDMD models for model predictive control~\cite{korda2018linear,arbabi2018data,korda2020optimal,kaiser2021data}, with recent extensions including incremental updates~\cite{calderon2021koopman} and Koopman autoencoders~\cite{retchin2023koopman}.  
Importantly, these approaches do not have the same convergence guarantees as the herein introduced KARL.
Further, MPC is an extremely robust control architecture that blends model-based feedforward control with sensor-based feedback control.  
As a result, MPC is often able to achieve robust performance, even with a relatively poor model, as discussed in~\cite{Kaiser2018prsa}.  
It is known that eDMD models often result in spurious eigenvalues and may perform badly with system noise and without tuning, so it is likely that the early Koopman-MPC work was largely exploiting the robustness of MPC rather than the model performance of eDMD.  
In contrast, KARL approaches rely on having a highly accurate model of the dynamics through the Koopman operator, putting a more more stringent requirement on the model actually predicting the dynamics.

\subsection{Contributions}
\label{subsec:contributions}

\begin{itemize}
    \item \textbf{Introduction of two max-entropy RL algorithms using the Koopman operator} {soft value iteration}~\cite{hazan2019provably} and {soft actor-critic}~\cite{haarnoja2018soft,haarnoja2018softapps}. We refer to this approach broadly as {Koopman-Assisted Reinforcement Learning (KARL)} and to the two particular reformulated algorithms as \textbf{soft Koopman value iteration (SKVI)} and \textbf{soft actor Koopman-critic (SAKC)}.
    \item \textbf{Validation} of these newly-introduced algorithms \textbf{on four environments}: a linear system, the Lorenz system, fluid flow past a cylinder, and a double-well potential with non-isotropic stochastic forcing.
    \item\textbf{State-of-the-Art Performance} is demonstrated across all four environments compared to state-of-the-art reinforcement learning SAC and classical control LQR baselines.
    \item The \textbf{Koopman tensor} is constructed in a multiplicatively separable dictionary space on states and controls, respectively. This approach extends and generalizes previous attempts to include actuation and control within the Koopman framework. Additionally, the present work extends Koopman for control to the context of Markov decision processes (MDPs).
    \item \textbf{Extensive ablation analyses} for SKVI and SAKC to discern the influence of the dictionary design choices in the construction of the Koopman tensor, and its performance.
\end{itemize}

\section{Background}
\label{Background}

Here, we discuss relevant background theory and algorithms. We review Koopman operator theory, discuss the reformulation of MDPs, and build intuition that will be used to develop the Koopman tensor in the next section.

\subsection{Koopman Operator Theory}
\label{subsec:koopman-operator-theory}

Nonlinearity is one of the central challenges in data-driven modeling and control, including modern RL architectures. Koopman operator theory~\cite{koopman1931hamiltonian,koopman1932dynamical,Mezic2004physicad,mezic2005spectral,Budivsic2012chaos,Mezic2013arfm,Brunton2022siamreview} provides an alternative perspective, in which nonlinear dynamics become linear when lifted to an infinite-dimensional Hilbert space of measurement functions of the system. This is consistent with other fields in machine learning, where lifting is known to linearize and simplify otherwise challenging tasks~\cite{Brunton2022book}. There are several forms a dynamical system may take to describe the evolution of a state $x$ in time.  A deterministic, autonomous system may be written in discrete-time as
\begin{align}
    x' = {F} (x),
\end{align}
or in continuous-time as
\begin{align}
    \frac{d}{dt}{x}(t) = {f}(x(t)).
\end{align}
Every continuous-time system induces a discrete-time system via the flow map operator
\begin{align}
    x(t+\tau) = {F}_{\tau} (x(t)) = x(t) + \int_{t}^{t+\tau} {f}(x(s)) ds.
\end{align}
The Koopman operator provides an alternative perspective. Formally, we consider real-valued vector measurement functions $g \colon M\to \mathbb{R}$, which are themselves elements of an infinite-dimensional Hilbert space and where $M$ is a manifold. Typically this manifold is taken to be $L^\infty(\mathcal{X})$ where $\mathcal{X}\subset \mathbb{R}^d$, where $d$ is the dimension of the embedding space. Often, these functions $g$ are called observables. The Koopman operator $\mathcal{K}:L^\infty(\mathcal{X})\to L^\infty(\mathcal{X})$, and its (infinitesimal) generator $\mathcal{L}$, are infinite-dimensional linear operators that act on observables $g$, in deterministic systems, as:
\begin{subequations}
    \begin{align}
        \mathcal{K} g &= g \circ F    \\
        \mathcal{L} g &= f \cdot \nabla g.
    \end{align}    
\end{subequations}
The Koopman generator $\mathcal{L}$ has the following relationship with the Koopman operator:
\begin{align}
    \mathcal{L}g = \lim_{t\to 0}\frac{\mathcal{K}g-g}{t} = \lim_{t\to 0}\frac{g\circ F -g}{t}.
\end{align} 
Koopman operator theory applies more broadly to any Markov process, although here we only give a simple example of a continuous-time stochastic system, the stochastic differential equation form for It\^{o}-diffusion processes:
\begin{align}
    dX(t) = \mu(X(t))dt + \sigma(X(t))dW(t)
\end{align}
where ${W(t)}$ is a Wiener process (i.e. standard Brownian motion). For stochastic systems, the definition of the Koopman operator is generalized to be defined as:
\begin{align}
    \mathcal{K} g &= \mathbb{E}(g(X)|X_0=\cdot)    \\
    \mathcal{L}g & = \lim_{t\to 0}\frac{\mathcal{K}g-g}{t},
\end{align}
where $\{X\}$ denotes the stochastic process representing the state over time. The Koopman operator advances measurement functions $g$ along the path of the trajectory $x$ as:
\begin{equation}
    \mathcal{K_\tau}g(x_t) := g(F_\tau(x_t)) =  g(x_{t+\tau}),
\end{equation}
where $F$ is the single-step flow map or law of motion. More generally in stochastic autonomous systems, it is defined as the conditional forecast operator:
\begin{align}
    \mathcal{K}_{\tau} g(x_t) &= \mathbb{E}\left(g(X_{t+\tau})|X_t=x_t\right).
\end{align}
Note that it is conventional, especially in discrete time, to denote $\mathcal{K} := \mathcal{K}_1 $ which we will be using exclusively below. Because the Koopman operator is infinite-dimensional, it is not immediately obvious that it is a computationally tractable alternative to working with the original finite-dimensional nonlinear dynamics. Much effort has gone into finding finite-dimensional Koopman invariant subspaces of measurements that behave linearly and also provide a useful basis in which to expand and approximate quantities of interest. These Koopman invariant subspaces are generically spanned by eigenfunctions of the Koopman operator.
\begin{remark}[Basis Functions of the Koopman Operator]
    The main insight from this operator theoretic view of dynamics is that a finite set of basis functions may be found to characterize observables of the state as long as the {Koopman operator has a finite point spectrum rather than a continuous spectrum}. This perspective is crucial below as we lift our coordinate spaces to a finite set of basis functions, typically called {dictionary functions}, which we denote by $\{\phi_i\}$.
\end{remark}

\noindent The Koopman operator has been widely studied and extended~\cite{Budivsic2012physd,Lan2013physd,Brunton2016plosone,korda2018convergence,mardt2020deep,brunton2021modern,colbrook2023beyond,colbrook2023mpedmd,colbrook2024rigorous}, for example using time delay coordinates as a universal embedding coordinate system~~\cite{Brunton2017natcomm,hirsh2021structured}. Similarly, the DMD algorithm has been widely expanded and extended to include several technical innovations~\cite{kunert2019extracting,hirsh2020centering,herrmann2021data,baddoo2023physics,hirsh2020data,azencot2019consistent,Hemati2017tcfd,Dawson2016ef,Askham2018siads} and has been applied to several applied domains, including fluid mechanics~\cite{Taira2017aiaa}, neuroscience~\cite{Brunton2014jnm}, and chemical kinetics~\cite{noe2013variational,nuske2014jctc,nuske2016variational,wu2017variational}. 

\noindent Controlled deterministic, discrete-time systems can similarly be written as: 
\begin{align}
    x' = {F} (x, u),
\end{align}
and controlled continuous-time systems can be written as:
\begin{align}
    \frac{d}{dt}{x}(t) = {f}(x(t), u(t)).
\end{align}
Discrete-time stochastic system with control can be expressed as
\begin{align}
    x' = {F} (x, u) + \sigma(x, u)\varepsilon'
\end{align}
where $\{\varepsilon\}$ is a white noise process ({$\varepsilon \sim \mathcal{N}(0,1)$}). Koopman has already been partially extended to these controlled dynamical systems~\cite{proctor2016dynamic,proctor2018generalizing}, and we will continue this extension with the Koopman tensor in the next section.  

\subsection{MDPs and Bellman's Equation}

Below, we assume an infinite horizon MDP setting for the agent's objective. We assume that the agent follows a policy $\pi(u|x)$, which is the probability of taking action $u$ given state $x$. In this case, and in discrete time, the $\pi$-value function takes the form:
\begin{align}
    V^\pi(x) = \mathbb{E}\left[\left. \sum_{k=0}^\infty-\gamma^k c(x_k,u_k)\right|\pi, x_0=x\right],
\end{align}
where $\gamma\in [0,1]$ represents the discount rate and we express rewards in terms of negative costs $r(x,u) = -c(x,u)$ as we will use a quadratic cost function as in the standard setting LQR below.
\begin{remark}[Finite-Horizon MDPs and the Time-Inhomogenous Koopman Operator]
    The finite horizon MDP can also be transformed using the Koopman operator, however, in that case, the operator will not be time-homogeneous and will depend not only on action, but also on point in time. For an excellent in-depth discussion of discrete-time MDPs (both finite and infinite horizon) and their place in RL, see~\cite{agarwal2022reinforcementlearning}.
\end{remark}

\noindent The agent's optimal value function is expressed as follows:
\begin{align}
    V^*(x) = \max_\pi V^\pi (x).
\end{align}
We use the Bellman equation (or Hamilton-Jacobi-Bellman equation in continuous time) to recursively characterize the above optimal value function. The discrete Bellman optimality equation is:
\begin{subequations}
    \label{Eq:BellmanKoopman}
    \begin{align}
        V(x) &= \max_{\pi} \mathbb{E}_{u\sim \pi(\cdot|x)}\left[-c(x,u) + \gamma \mathbb{E}_{x'\sim p(\cdot|x,u)}[V(x')]\right]\\
             &= \max_{\pi} \mathbb{E}_{u\sim \pi(\cdot|x)}\left[-c(x,u) + \gamma \mathcal{K}^uV(x)\right]\\
             &= \max_{\pi} \mathbb{E}_{u\sim \pi(\cdot|x)}\left[-c(x,u) + \gamma w^T{K}^u\phi(x)\right],
    \end{align}
\end{subequations}
where $\phi: \mathcal{X}\to \mathbb{R}^{d_x}$ is the dictionary vector function $\phi=(\phi_1,\cdots,\phi_{d_x})$ that serves as a basis for the Koopman operator applied to the value function. Note that the Koopman operator applies element-wise when applied to vector functions.

\subsubsection{Continuous-Time Formulation}

In continuous time, the value function from following policy $\pi$ takes the form:
\begin{align}
    V^\pi(x) = \mathbb{E}\left[\left. \int_{t=0}^\infty -e^{-\rho t}c(x_t,u_t)dt\right|\pi, x_0=x\right].
\end{align}
The agent's optimal value function is again defined as:
\begin{align}
    V^*(x) = \max_\pi V^\pi (x).
\end{align}
We now use HJB to recursively characterize the optimal value function:
\begin{align}
    V(x) &= \frac{1}{\rho}\max_{\pi_t} \mathbb{E}_{u\sim \pi_t(\cdot|x)}\left[-c(x,u) +  \mathcal{L}^uV(x)\right].
\end{align}
where $\mathcal{L}$ is the generator of the Koopman operator as defined above.
\begin{remark}[Closed Form Operator Representation of $V^\pi$]
    For a deterministic system, the value function may be recovered from the Koopman operator $\mathcal{K}^\pi$, given a fixed policy $\pi$: 
    \begin{align}
        V^\pi(x) = -c(x,u)+\gamma V^\pi(x') = -c(x,u)+\gamma \mathcal{K}^\pi V^\pi(x) \quad\Longrightarrow\quad V^\pi(x) = -\left({I}-\gamma \mathcal{K}^\pi\right)^{-1}c(x,u).
    \end{align}
\end{remark}

\section{Koopman with Control from a Policy Perspective}

There have been several attempts to develop extensions of Koopman operator theory to incorporate control~\cite{proctor2016dynamic,proctor2018generalizing,kaiser2021data}. Here, we develop some simple results that enable transformations between different Koopman with control representations.  Figure~\ref{fig:KoopmanControl} shows a basic schematic that will be useful for the discussion.  Note that this section only deals with deterministic systems, although the arguments may be readily extended to the stochastic setting. If the control action $u$ is a constant for the entire trajectory, then the controlled system may be viewed as an autonomous system parameterized by $u$, and we may define ${F}^u$ and $\mathcal{K}^u$ as:

\begin{figure}[t]
    \centering
    \includegraphics[width=\textwidth]{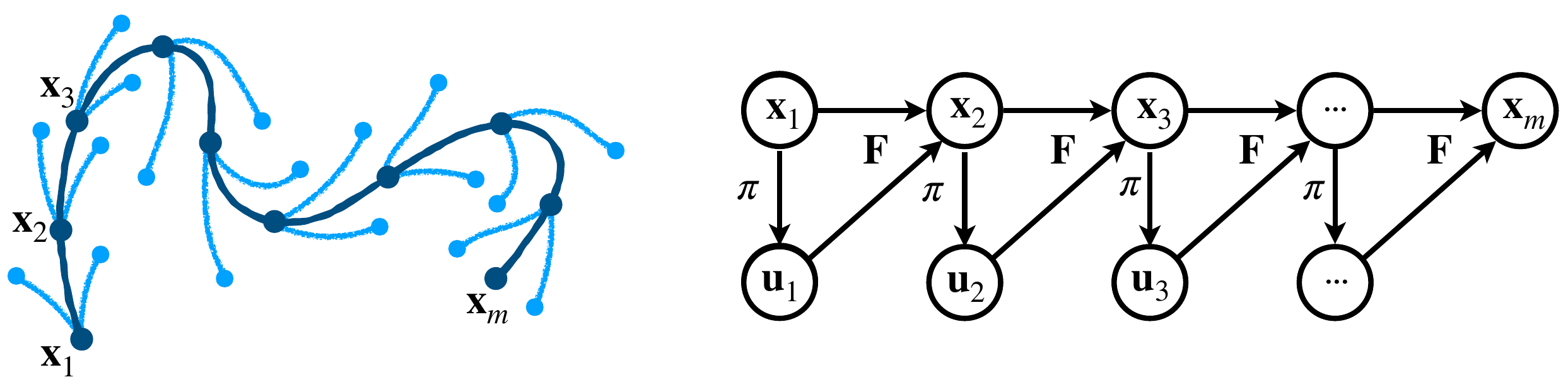}
    \caption{
        Schematic of Koopman with control. (left) A nominal trajectory is shown for a given policy $\pi$, with alternative branches for different control actions.  (right) A diagram showing the state and control sequences, as they relate to the policy $\pi$ and dynamics ${F}$.
    }
    \label{fig:KoopmanControl}
\end{figure}

\begin{subequations}
    \begin{align}
        x' &= {F}(x,u) := {F}^u(x)\\
        g(x') & = g({F}^u(x)):=\mathcal{K}^ug(x).
    \end{align}
\end{subequations}
Similarly, for a given fixed policy $\pi(u,x)$, we may define the autonomous systems ${F}^\pi$ and $\mathcal{K}^\pi$ as:
\begin{subequations}
    \begin{align}
        x' &= {F}(x,\pi(x)) := {F}^\pi(x)\\
        g(x') & = g({F}^\pi(x)):=\mathcal{K}^\pi g(x).
    \end{align}
\end{subequations}
In both cases, for a fixed control $u$ or a fixed control law $u=\pi(x)$, the dynamics may be viewed as autonomous and standard Koopman operator techniques may be used to approximate $\mathcal{K}^u$ and $\mathcal{K}^\pi$. If $\pi(x)=u$ at a specific point $x$, then the following
\begin{align}
    g(x') = \mathcal{K}^u(x)
\end{align}
and
\begin{align}
    g(x') = \mathcal{K}^\pi(x)
\end{align}
are equivalent at the point $x$, although they are not necessarily equivalent at other points. This has implications for representations in a basis $\{\phi_j(x)\}_{j=1}^n$. If $g(x)={w}^T\boldsymbol{\phi}(x)$ then
\begin{align}
    \mathcal{K}^u g(x) = {w}^{\top}{K}^u\phi(x)
\end{align}
and 
\begin{align}
    \mathcal{K}^\pi g(x) = {w}^{\top}{K}^\pi\phi(x)
\end{align}
must be equal at the point $x$. Thus, the following must be true at $x$:
\begin{align}
    {w}^{\top}{K}^u\phi(x)={w}^{\top}{K}^\pi\phi(x).
\end{align}

\section{Koopman-Assisted Reinforcement Learning (KARL)}
\label{KARL}

In this section, we leverage our reformulation of MDPs via the Koopman operator in equation~\eqref{Eq:BellmanKoopman} to develop two new maximum entropy RL algorithms: {Soft Koopman Value Iteration} and {Soft Actor Koopman-Critic}. We subsequently describe the Koopman tensor formulation that is employed to develop and implement Koopman-assisted reinforcement learning algorithms in practice.

\subsection{Maximum Entropy Koopman RL Algorithms}

To demonstrate the effectiveness of how the Koopman operator can be used in RL we follow a popular strand of RL literature and add an {entropy penalty}, $\alpha \cdot \ln (\pi)$, to the cost function to encourage exploration of the environment.

\begin{figure}[t]
    \centering
    \includegraphics[width=\textwidth]{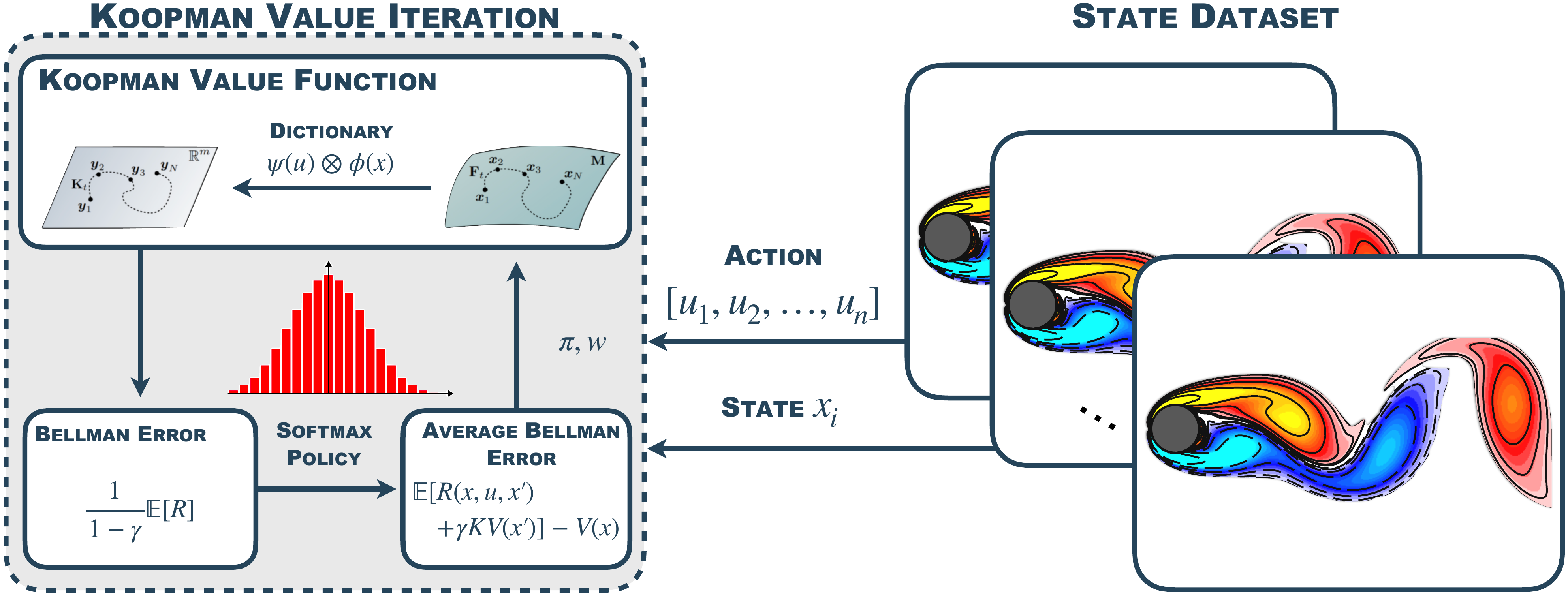}
    \caption{
        Soft Koopman Value Iteration, a Koopman variant of the widely used value iteration algorithm. In the Koopman value iteration, The set of states $x_{i}$, under a sequence of actions $\{u_{j} \}_{j=1,\ldots,n}$, are lifted onto the vector space to advance the dynamics with the Koopman operator linearly. The action policy is then learned in this new space.
    }
    \label{fig:koopman-value-iteration}
\end{figure}

\subsubsection{Soft Koopman Value Iteration}

The proposed soft Koopman value iteration approach is shown in Fig.~\ref{fig:koopman-value-iteration}. In addition to assuming that we have a finite-dimensional approximation of the Koopman operator, we also assume that the optimal value function $V^\star(x)$ can be written as a linear combination of basis functions. In other words, there exists a $w^\star\in\mathbb{R}^d$, such that $V^\star(x) = (w^\star)^{\top} \phi(x)$. Given a $w\in \mathbb{R}^{d}$, we can express the entropy regularized Bellman error as follows. For any $x$ the Bellman error is:
\begin{equation}
    \label{eq:be}
    w^{\top}\phi(x) - \min_{\pi: \mathcal{X}\mapsto \Delta(\mathcal{U})} \mathbb{E}_{u\sim \pi(x)} \left[c(x,u) + \alpha\ln\pi(u | x) + w^{\top} K^u \phi(x) \right].
\end{equation}
Thanks to the entropy regularization, given a $w$, we can express the optimal form of $\pi$ as follows:
\begin{equation}
    \label{eq:softmax_policy}
    \pi(u | x) = \exp\left( - \left( c(x,u) + w^{\top}K^u \phi(x)   \right)/\alpha  \right) / Z_x,
\end{equation}
where $Z_x$ is the normalizing constant that only depends on $x$ and makes $\pi(\cdot|x)$ a proper probability distribution with its probabilities summing to 1. Note that $\pi$ depends on $w$. Converting this into an iterative procedure to find the value function weights, $w'$, in terms of the previous weights, $w$, the {average Bellman error (ABE)} over the data can be expressed as:
\begin{equation}
    \min_{w': \|w\|_2 \leq W} \frac{1}{N}\sum_{i=1}^N \left(w'^{\top}\phi(x) - \min_{\pi(\cdot | x)} \mathbb{E}_{u\sim \pi(x)} \left[ c(x,u) + \alpha\ln\pi(u|x) +  w^{\top} K^u   \phi(x) \right] \right)^2.
\end{equation}
This is a canonical ordinary least squares (OLS) problem that can be solved explicitly for $w'$. We repeat this procedure until the ABE is small or until convergence. Unfortunately, given finite data, it is possible for a suboptimal value function to satisfy the Bellman equation exactly~\cite{fujimoto2022should}. Thus, we must be careful when using the Bellman error as a training objective for RL agents.

\begin{algorithm}[t]
    \caption{Optimal Policy via Average Bellman Error (ABE) Minimization -- Soft Koopman Value Iteration (SKVI)}
    \begin{algorithmic}[1]
        \REQUIRE Confidence parameter $\epsilon$, reward function $r$, feature maps $\phi: \mathcal{X} \to \mathbb{R}^{d_x}$ and $\psi: \mathcal{U} \to \mathbb{R}^{d_u}$, datasets $\mathcal{D}_x$ and $\mathcal{D}_u$, initial weights $w_0$
        \STATE Let $ABE$ denote the objective of the minimization problem \eqref{eq:be}
        \STATE Let $\pi_0^*(u|x)$ be the optimal policy given in \eqref{eq:softmax_policy} evaluated at $w_0$
        \STATE $ABE(w_i, \pi_i) \gets ABE(w_0, \pi_0^*)$
        \WHILE {$ABE(w_i, \pi_i) > \epsilon$}
            \STATE Sample $x_i \sim \mathcal{D}_x$
            \STATE Compute costs, log probabilities, and value function outputs for all $(x_i, u)$ pairs, $u \in \mathcal D_u$
            \STATE $w_{i+1} \gets$ solution to Equation \eqref{eq:be}
            \STATE $w_{i+1} \mapsto \pi_{i+1}^*(u|x)$
            \STATE $ABE(w_i, \pi_i) \gets ABE(w_{i+1},\pi_{i+1}^*(u|x))$
        \ENDWHILE
    \end{algorithmic}
\end{algorithm}

\subsubsection{Soft Actor Koopman-Critic}
\label{sec:sakc}

Here, we outline how we modify the Soft Actor-Critic framework~\cite{haarnoja2018soft} to restrict the search space by incorporating information from the Koopman operator. An illustration of the Soft Actor Koopman-Critic can be seen in Fig.~\ref{fig:soft-actor-koopman-critic}. Using the same loss functions and similar notation to that of the SAC paper~\cite{haarnoja2018soft}, we first specify the soft value function loss:
\begin{equation}
    J_V(w) = \mathbb{E}_{x \sim \mathcal{D}} \left[ \frac{1}{2}\left(V_w(x) - \mathbb{E}_{u\sim \pi_\phi}\left[Q_\theta(x, u) - \alpha\ln \pi_\nu(u|x)\right]\right)^2\right]. \label{eq:soft_value_loss}
\end{equation}
The additional specification that is imposed in the Koopman reinforcement learning framework would be a restriction around the specifications of $V_w(x)$ and $Q_\theta(x, u)$:
\begin{equation}
    V_w(x) = w^T \phi(x),
\end{equation}
where $w$ is a vector of coefficients for the dictionary functions. Next, we show how the loss function for the quality function $Q$ changes:
\begin{equation}
    J_Q(\theta) = \mathbb{E}_{(x, u)\sim \mathcal{D}}\left[\frac{1}{2}\left(Q_\theta(x, u) - \widehat Q(x, u)\right)\right],
\end{equation}
where the target $Q$-function incorporates the Koopman operator and is defined as:
\begin{align}
    \widehat Q(x, u) &= r(x, u) + \gamma \mathbb{E}_{x\sim p(\cdot|x,u)}\left[V_{\bar w}(x)\right]\notag
    \\
    &= r(x, u) +  \gamma \bar w^T K^{u}\phi(x),
\end{align}
and where $\mathcal{K}$ represents the infinite-dimensional Koopman for a fixed action $u$ and $K^{u}$ represents the finite-dimensional approximation of the Koopman operator in the state-dictionary space. Finally, the loss function for the policy does not change and is given by:
\begin{equation}
    J_\pi (\nu) = \mathbb{E}_{x\sim \mathcal{D}}\left[D_{KL}\left(\pi_\nu(\cdot|x) \left \| \frac{\exp(Q_\theta(x,\cdot))}{Z_\theta(x)}\right) \right]. \right .
\end{equation}
After these adjustments, the algorithm remains the same as in SAC and is given by Algorithm~\ref{alg:sac}.
\begin{algorithm}
    \caption{Soft Actor Koopman-Critic}
    \label{alg:sac}
    \begin{algorithmic}[1]
    \REQUIRE Initial parameter vectors $\nu$, $w$, $\theta$, $\bar{w}$ 
    \FOR{$\text{each iteration}$}
        \FOR{$\text{each environment step}$}
            \STATE $u {\sim} \pi_{\nu}(u | x)$
            \STATE $x' \sim p(x' | x, u)$
            \STATE $\mathcal{D} \leftarrow{ \mathcal{D} \cup \{ (x, u, r(x, u), x' \} }$
        \ENDFOR
        \FOR{$\text{each gradient step}$}
            \STATE $w \leftarrow{ w - \lambda_{V} \hat{ \nabla}_{\nu} J_V(w) } $
            \STATE $\theta_{i} \leftarrow{ \theta_{i} - \lambda_{Q} \hat{\nabla}_{\theta_{i}} J_{Q(\theta_{i})} \text{ for } i \in \{ 1, 2 \} }$
            \STATE $\nu \leftarrow{ \nu - \lambda_\pi \hat{\nabla}_\nu J_\pi(\nu) }$
            \STATE $\bar{w} \leftarrow{\tau w + (1 - \tau)\bar{w} }$
        \ENDFOR
    \ENDFOR
    \end{algorithmic}
\end{algorithm}

\subsection{Koopman Tensor Formulation of Controlled Dynamics}
\label{sec:koopman-tensor-formulation}

We now focus on how to advance the basis functions via the Koopman operator, given the current state and action. That is, we would like to find the mapping $(x,u) \mapsto \phi(x')$. We also impose that the dictionary functions approximately span a finite-dimensional Koopman-invariant subspace of the value function for each $u$, so that there exists a matrix $K^u\in \mathbb{R}^{d_x\times d_x}$ such that $K^u\phi(x) = \phi(x')$. The goal is to construct the Koopman matrix $K^u$ for every action $u\in\mathcal{U}$ given state and action trajectory data and a state dictionary space $\phi$. To do so, we follow an approach similar to that described in SINDYc \cite{brunton2016sparse,Kaiser2018prsa} where a dictionary space on states and actions is used to predict the next state, that is $\Theta(x,u)\mapsto x'$. There are 2 important differences in our approach. First, we are not trying to predict the next state $x'$, but rather the next dictionary function value $\phi(x')$. Second, to respect the fact that the state dictionary space spans a Koopman invariant subspace for every $u$, it must be that the state-action dictionary is separable in state and action. Modeling the state-action dictionary space as multiplicatively separable as $\psi(u)\otimes \phi(x)$ and subsequently assuming that there exists a linear mapping $\psi(u)\otimes\phi(x)\mapsto \phi(x')$ allows us to construct a Koopman matrix $K^u$ for every action $u$. In what follows, we formally describe how constructing a tensor from the linear operator $\psi\otimes\phi \mapsto \phi$ allows us to calculate the matrix $K^u$ for any given u.

\begin{figure}
    \centering
    \includegraphics[width=\textwidth]{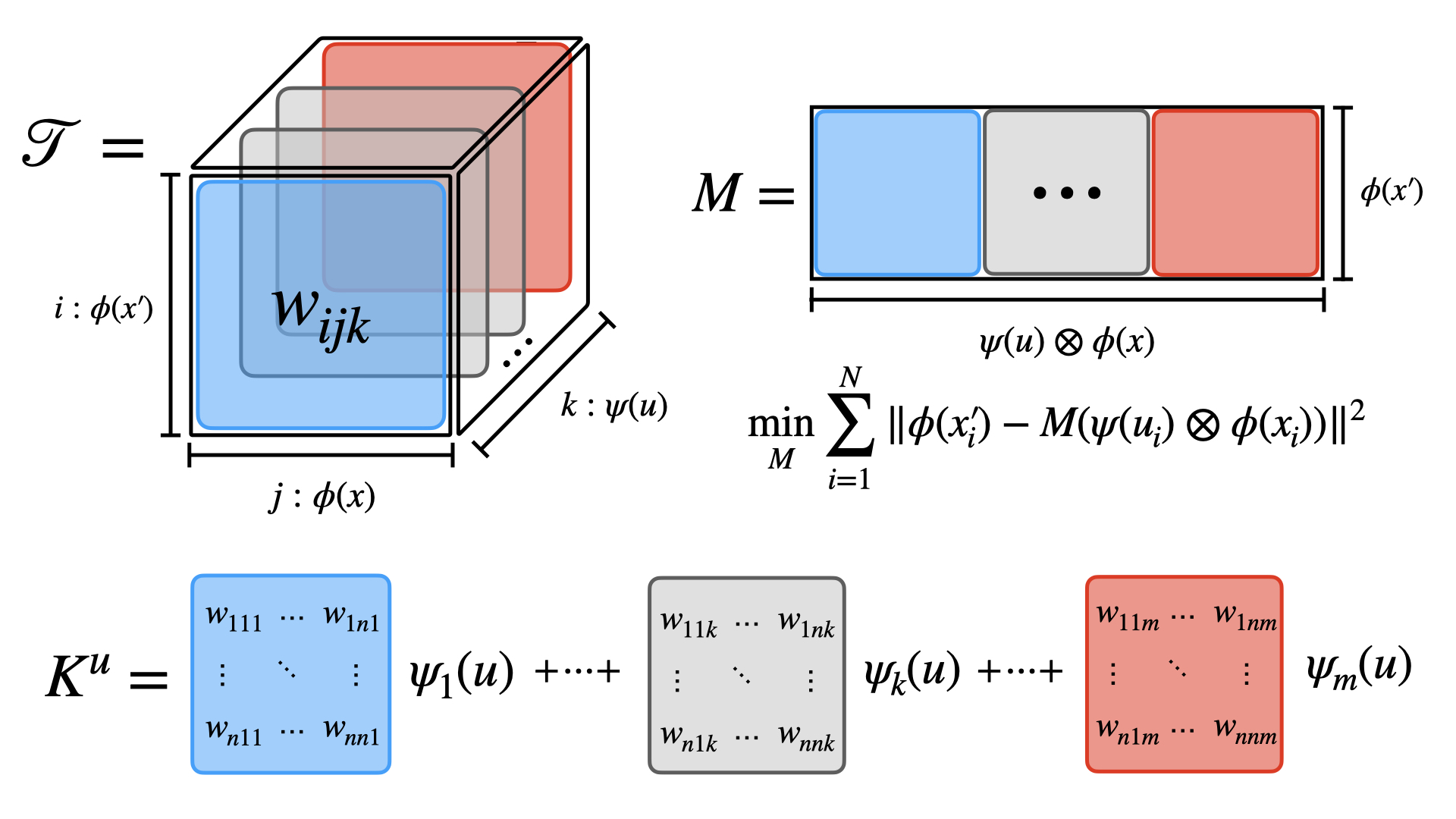}
    \caption{
        Construction of action-dependent Koopman operators $K^u$ from the Koopman tensor $\mathscr{T}_K$. Colors match along the $k$ index (depth of tensor box). Each of the matrix slices is then weighted according to the $\psi$ dictionary elements to construct the control-dependent Koopman operator $K^u$.
    }
    \label{fig:koopman_tensor}
\end{figure}

\begin{algorithm}
    \caption{Koopman Tensor Estimation} 
    \begin{algorithmic}[1]
	\REQUIRE State feature map $\phi: \mathcal{X} \to \mathbb{R}^{d_x}$, control feature map $\psi: \mathcal{U} \to \mathbb{R}^{d_u}$, and a sample $\{(x_i,u_i)\}_{i=0}^N$\;\\
	\STATE Solve for $\widehat{M}$ as in Equation \ref{eq:fit-koopman-tensor}\\
        \STATE Convert $M$ into $\mathscr{T}_K$ tensor using Fortran-style reshaping
    \end{algorithmic} 
\end{algorithm}

\noindent Denote $\phi: \mathcal{X}\mapsto \mathbb{R}^{d_x}$ as the feature mapping for the state (each coordinate of $\phi$ is an observable function), and $\psi:\mathcal{U} \mapsto \mathbb{R}^{d_u}$ as the feature mapping for control actions. We seek a finite-dimensional approximation of the Koopman operator $\mathcal{K}^{u}$ for all $u \in \mathcal{U}$. Denote $T_K \in \mathbb{R}^{d_x\times d_x \times d_u}$ as a 3D tensor as shown in Fig.~\ref{fig:koopman_tensor}. For any $u$, define $K^{u}\in\mathbb{R}^{d_x\times d_x}$ as follows: $K^{u}[i,j] = \sum_{z = 1}^d T_K(i, j, z) \psi(u)[z]$. Namely, $K^u$ is the result of the tensor vector product along the third dimension of $T_K$ and $K^u$ serves as the finite-dimensional approximation of Koopman operator $\mathcal{K}^u$. We learn $T_K$ to minimize the error in advancing the basis functions $\phi$, averaged over the data:
\begin{equation}
    \min_{T_K} \sum_{i=1}^N \left\|  K^{u_i} \phi(x_i) - \phi(x_i')  \right\|^2.
\end{equation}

\noindent We can rewrite the above objective so it becomes a regular multi-variate linear regression problem, rearrange $T_K$ as a 2-dimensional matrix in $\mathbb{R}^{d_x\times d_x \cdot d_u}$. Denote $M \in \mathbb{R}^{d_x\times d_x \cdot d_u}$, where $M[i, :] \in \mathbb{R}^{d_x \cdot d_u}$ is the vector from stacking the columns of the 2D matrix $T_K[i, :, :]$, as shown in Fig.~\ref{fig:koopman_tensor}. Denote $ \psi(u)\otimes \phi(x)\in\mathbb{R}^{d_x \cdot d_u}$ as the Kronecker product. Thus we have:
\begin{equation}
    K^{u} \phi(x) = M ( \psi(u)\otimes \phi(x) ).
\end{equation}
Therefore, the optimization problem becomes a regular linear regression:
\begin{equation}
    \label{eq:fit-koopman-tensor}
    \min_{M} \sum_{i=1}^N \left\|  M \left( \psi(u)\otimes  \phi(x_i)\right) - \phi(x_i')  \right\|^2.
\end{equation}

\noindent Once we compute $M$, we can convert back to a standard Koopman operator for any $u \in \mathcal{U}$ by reshaping $M$ back to the 3D tensor $T_K$. Then the $d_x\times d_x$ finite dimensional Koopman operator approximation is again $K^{u}$  for any $u \in \mathcal{U}$ as seen in the summation in Fig.~\ref{fig:koopman_tensor}. Note that the above formulation also works for a discrete control set $\mathcal{U}$, which could improve sample efficiency as opposed to learning independent Koopman operators one for each discrete control.

\section{Evaluation}
\label{Evaluation}

\begin{figure}[t]
    \centering
    \includegraphics[width=.9\textwidth]{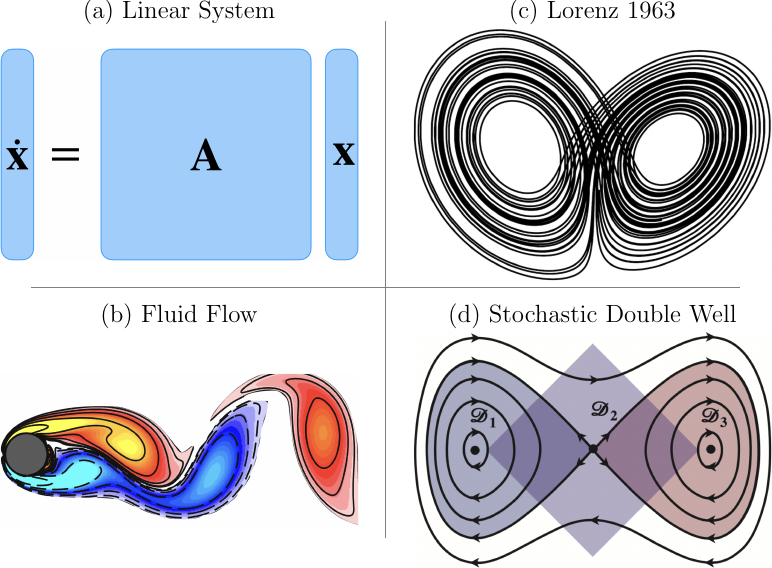}
    \caption{
        Four benchmark problems examined: \textbf{(a)} a simple linear system; \textbf{(b)} a Lorenz 1963 model; \textbf{(c)} an incompressible fluid flow past a cylinder at Reynolds number 100; and \textbf{(d)} a double-well potential with non-isotropic stochastic forcing.
     }
    \label{fig:evaluation}
\end{figure}
To evaluate our KARL algorithms, we implemented four benchmark environments widely used in the classical dynamics and control literature.  These environments are shown in Fig.~\ref{fig:evaluation}. In addition, we implemented a classic control baseline. For this purpose, we implemented the linear quadratic regulator and utilized the SAC implementation of CleanRL~\cite{huang2022cleanrl}. We will first detail the environments before going into the design of the evaluation experiments.

\paragraph*{Linear System.} 
This environment is the general form of a controlled, linear system. The linear quadratic regulator (LQR) is optimal for this system. The discrete-time dynamics for the linear system are given by the following, where $A$, and $B$ are matrices:
\begin{equation}
    F(x, u) = Ax + Bu,
\end{equation}

\paragraph*{Fluid Flow.} 
Approximating the fluid flow past a cylinder, we build on the reduced-order model developed by Noack et al.~\cite{Noack2003jfm}, used for the testing of Koopman modeling algorithms such as in Lusch et al.~\cite{lusch2018deep} previously. The continuous-time dynamics are:
\begin{equation}
    {f}(x, u) = \begin{bmatrix}
        \mu x_0 - \omega x_1 + A x_0 x_2\\
        \omega x_0 + \mu x_1 + A x_1 x_2 + u\\
        -\lambda ( x_2 - x_0^2 - x_1^2 )
    \end{bmatrix},
\end{equation}
where $\mu = 0.1$, $\omega = 1.0$, $A = -0.1$, and $\lambda = 1$. The states $x_{0}$ and $x_{1}$ represent the most energetic proper orthogonal decomposition modes for the flow, and the third state $x_{2}$ represents the shift mode, used for capturing the relevant transients.

\paragraph*{Lorenz 1963.} 
A canonical chaotic dynamical system used to evaluate Koopman-based approaches due to its continuous eigenvalue spectrum~\cite{Brunton2017natcomm}. The continuous-time dynamics are governed by the following system, where $\sigma = 10$, $\rho = 28$, and $\beta = \frac{8}{3}$:

\begin{equation}
    {f}(x, u) = \begin{bmatrix}
        \sigma ( x_1 - x_0 ) + u\\
        ( \rho - x_2 ) x_0 - x_1\\
        x_0 x_1 - \beta x_2
    \end{bmatrix},
\end{equation}

\paragraph*{Stochastic Double Well.} 
Considering the dynamics of a stochastically forced particle in a double well potential, the dynamics are given by the following equation system:

\begin{equation}
    {f}(x, u) =
    \begin{bmatrix}
        4x_0 - 4 x_0^3 + u\\
        -2x_1 + u
    \end{bmatrix}
    +
    \begin{bmatrix}
        0.7 & x_0\\
        0 & 0.5
    \end{bmatrix}
    \begin{bmatrix}
        v_0 \sim \mathcal{N}(0, 1)\\
        v_1 \sim \mathcal{N}(0, 1)
    \end{bmatrix}.
\end{equation}

\paragraph*{Cost Function.}
For all policies, and for all systems we use the LQR cost function, which is defined by:
\begin{equation}
    c(x, u) = (x-x_e)^{\top} Q (x-x_e) + u^{\top} R u,
\end{equation}
where $x_e$ is the reference point we wish to stabilize, and $Q$ and $R$ are positive semi-definite matrices. For all environments except Lorenz, $x_e$ is the origin and for Lorenz, it is the critical point:
\begin{equation}
    x_e =
    \begin{bmatrix}
        x_{1e} \\ x_{2e} \\ x_{3e}
    \end{bmatrix}
    =
    \begin{bmatrix}
        \sqrt{\beta ( \rho - 1 )}\\
        \sqrt{\beta ( \rho - 1 ) }\\
        \rho - 1
    \end{bmatrix}.
\end{equation}

\begin{figure}[t]
    \centering
    \includegraphics[width=\textwidth]{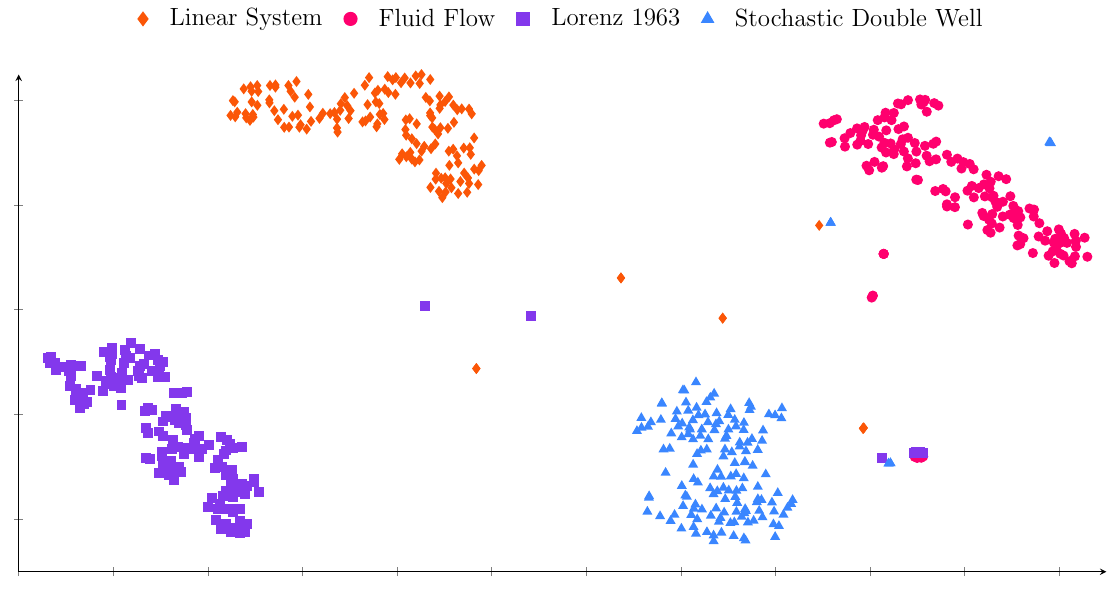}
    \caption{
        t-distributed stochastic neighbor embedding of the Koopman tensors projected onto a common basis. The Koopman tensors of the 4 dynamical systems show a clear separation into the distinct dynamics, as expected from a working Koopman tensor construction approach. 
    }
    \label{fig:koopman-t-sne}
\end{figure}

\paragraph*{Koopman Model Evaluation and Embedding.}
Before using for control, we verify our Koopman operator models for these for systems.  Specifically, we sample the space of potential Koopman operators for our example systems by varying the order of the dictionaries used for the state-, and action-space. These operators are subsequently projected onto a joint space of dictionaries, before collapsing them onto a 2-dimensional manifold using the t-distributed stochastic neighbor embedding (t-SNE) approach~\cite{van2008visualizing}. Figure~\ref{fig:koopman-t-sne} shows clear clustering of the Koopman operators into the distinct dynamical systems, which allows us to conclude that our operator approach works as designed.

\subsection{Baseline Control and Learning Algorithms}

We follow the established baseline implementation of the Python Control Systems Library for the linear quadratic regulator (LQR)~\footnote{\href{https://github.com/python-control/python-control/blob/main/control/statefbk.py\#L299}{github.com/python-control/python-control/blob/main/control/statefbk.py\#L299}}, and the CleanRL library~\footnote{\href{https://github.com/vwxyzjn/cleanrl/blob/master/cleanrl/sac_continuous_action.py}{github.com/vwxyzjn/cleanrl/blob/master/cleanrl/sac\_continuous\_action.py}} implementation of the Soft Actor-Critic baselines for the other cases. We have two Soft Actor-Critic baselines, which are different implementations of the Soft Actor-Critic model. The first baseline is from the original paper~\cite{haarnoja2018soft} and explicitly makes use of function approximation for both $Q$ and $V$. We denote this algorithm by \textit{SAC (V)}. This baseline algorithm is the same as Algorithm~\ref{alg:sac}, except that the value function $V_w$ is approximated by a neural network rather than a linear combination of the dictionary functions. In this implementation, there are four neural networks: $V_w$, $Q_{\theta_1}$, $Q_{\theta_2}$, and $\pi_\nu$ with parameter vectors $w, \theta_1, \theta_2$ and $\nu$, respectively. The following implementation details apply:
\begin{itemize}
    \item The policy $\pi_\nu$ is assumed to be the normal density function with mean and standard deviation being represented by the policy neural network.
    \item All networks are fully connected feed-forward networks of 1 hidden layer with 256 neurons, each with ReLU activation functions.
    \item The standard deviation is "squashed" with a $\tanh$ transformation, and a convex combination transformation that maintains the log standard deviation of the policy between $2$, and $-5$.
    \item The learning rate of both the Q and V networks are taken to be $1$e-$3$, whereas the policy network learning rate is taken to be $3$e-$4$.
    \item The replay pool (or buffer) is of max size $1$e$6$, and is initialized with $5000$ interactions in the environment following the initial policy. After $5000$ steps, the policy starts to update and after $1$e$6$ total steps, a FIFO replacement scheme is followed for updating the replay buffer.
\end{itemize}

\noindent Finally, we use the standard implementation of SAC from \cite{haarnoja2018softapps} as a baseline, denoted as \textit{SAC (Q)}. This implementation removes the V-network and the V-target network is replaced with a Q-target network. All other implementation details remain the same as above and the full algorithm is included in appendix~\ref{appendix:rl_algorithm_definitions} for completeness.

\subsection{KARL Algorithm Implementation Specifics}

\paragraph*{Construction of the Koopman Tensor.} The Koopman tensor and its action-dependent feature space Koopman operator $K^u$ are used in both the SKVI, and SAKC algorithm. The dataset from which the Koopman tensor is constructed is comprised of $3$e$+4$ interactions with the environment under a random agent, unless indicated otherwise. This agent samples its actions from a uniform distribution with bounds dependent on the specific environment. The bounds were chosen according to the minimum and maximum actions taken by an LQR policy.

\paragraph*{Soft Koopman Value Iteration.} SKVI was implemented with a manual calculation of the average Bellman error given a set of system states, and the extraction of the softmax policy. For the discrete value iteration, we use the same dataset used for the construction of the Koopman tensor. The action space is discretized such that the number of actions evenly split the system's range.

\paragraph*{Soft Actor Koopman-Critic.} SAKC follows Algorithm \ref{alg:sac} with the cost and target functions as in section~\ref{sec:sakc}. The algorithm mirrors all specification details of SAC (V) with the exception of the representation of the value function, and its continuation value. Specifically, as discussed in section~\ref{sec:sakc}, the value function $V_w(x)$ is approximated in the feature space as $V_w(x) = w^T \phi(x)$ and the continuation value is represented with the Koopman operator in the state feature space as $\mathbb{E}_{x\sim p(\cdot|x,u)}\left[V_{w}(x)\right] = w^T K^{u}\phi(x)$. The learning rate on the parameter $w$ is set to $1$e-$3$.

\subsection{Design of Evaluation}
We run a set of four experiments to measure episodic returns on all environments, comparing to baselines from classic control theory and state-of-the-art reinforcement learning. To examine the performance of the two proposed KARL variants, we first calibrate the hyperparameters of the Koopman-assisted algorithms using Bayesian optimization~\cite{bayesoptpackage2014,gardner2014bayesian} with an Asynchronous Hyperband scheduler~\cite{li2020system} in Ray Tune~\cite{liaw2018tune}. The optimization procedure is afforded a budget of $50$ runs for this purpose on each environment for each of the two algorithms. The definition of the hyperparameter optimization search space is summarized in Tab.~\ref{eval:hpo_bounds}. The best found hyperparameter configurations for each environment are condensed in Tab.~\ref{tab:hyperparameter_configuration} with the optimized hyperparameters highlighted in pink. To accurately assess the performance of each algorithm on each of the four environments, we make sure to run them for 25 seeds to protect against outliers. Following plotting standards for reinforcement learning~\cite{agarwal2021deep}, we use the inter-quartile mean as aggregate statistic, combined with $95\%$ stratified bootstrap confidence intervals computed with $50,000$ samples afforded to the bootstrapping sampler. %
To further discern the impact of individual design choices on the performance of the \textit{Soft Koopman Value Iteration}, and the \textit{Soft Actor Koopman Critic}, we perform a set of two ablation studies. In the case of the Soft Koopman Value Iteration, we perform a grid-sampling where the number of actions performed by the algorithm, and the number of epochs the initial Koopman model is trained over are varied. The exact grid-sampling configuration is summarized in Tab.~\ref{eval:ablation_configuration}. For each individual grid point, the experiments are run 5 times with random seeds of which the inter-quartile mean is then taken.

\begin{table}[t]
    \centering
    \caption{
        Grid-sampling configuration of the ablation studies, where the indicated parameters are varied while the remaining parameters are to remain fixed on the templated default parameters.
    }
    \label{eval:ablation_configuration}        

\begin{tabular}{ccc}
        \toprule
        {\renewcommand{\arraystretch}{1}\begin{tabular}[c]{@{}c@{}}\textbf{\textbf{Hyperparameter}}\vspace{.0em}\end{tabular}} & {\renewcommand{\arraystretch}{1}\begin{tabular}[c]{@{}c@{}}\textbf{\textbf{Soft Koopman Value Iteration}}\vspace{.0em}\end{tabular}} & {\renewcommand{\arraystretch}{1}\begin{tabular}[c]{@{}c@{}}\textbf{\textbf{Soft Actor Koopman Critic}}\vspace{.0em}\end{tabular}}\\
        \midrule
        \multicolumn{3}{c}{\textit{Algorithm-specific Arguments}}\\
        \midrule
        \begin{tabular}[c]{@{}c@{}}\textbf{Number of}\\\textbf{Actions}\end{tabular} & $\{ 71, 81, 91, 101, 111, 121 \}$ & \\
        \greyrule
        \begin{tabular}[c]{@{}c@{}}\textbf{Number of}\\\textbf{Training Epochs}\end{tabular} & $\{ 75, 100, 125, 150, 175, 200 \}$ & \\
        \greyrule
        \begin{tabular}[c]{@{}c@{}}\textbf{Value Network}\\\textbf{Learning Rate}\end{tabular} & & $\{ 0.0001, 0.0005, 0.001, 0.005, 0.01, 0.05 \}$ \\
        \greyrule
        \begin{tabular}[c]{@{}c@{}}\textbf{Policy Network}\\\textbf{Learning Rate}\end{tabular} & & $\{ 0.0001, 0.0005, 0.001, 0.005, 0.01, 0.05 \}$ \\
        \bottomrule
\end{tabular}

\end{table}

\noindent To showcase the interpretability inherent to the presented Koopman-based approach, we furthermore extract the polynomial representation of the Koopman-constructed value function for the hyperparameter-optimized configurations of the Soft Koopman Value Iteration for each of the four environments.

\begin{table}[t]
    \centering
    \caption{
        Hyperparameter optimization search space used in the dialing in of the hyperparameters. The search spaces are centered on the default hyperparameter configurations of the \textit{Soft Koopman Value Iteration}, and the \textit{Soft Actor Koopman-Critic}. The default values can be found in the single-file implementations of the algorithms in the Python package. Parameters are broken down into algorithm-specific arguments, and Koopman-specific arguments.
    }
    \label{eval:hpo_bounds}
    \resizebox{\textwidth}{!}{

\begin{tabular}{ccc}
        \toprule
        {\renewcommand{\arraystretch}{1}\begin{tabular}[c]{@{}c@{}}\textbf{\textbf{Hyperparameter}}\vspace{.0em}\end{tabular}} & {\renewcommand{\arraystretch}{1}\begin{tabular}[c]{@{}c@{}}\textbf{\textbf{Soft Koopman Value Iteration}}\vspace{.0em}\end{tabular}} & {\renewcommand{\arraystretch}{1}\begin{tabular}[c]{@{}c@{}}\textbf{\textbf{Soft Actor Koopman Critic}}\vspace{.0em}\end{tabular}}\\
        \midrule
        \multicolumn{3}{c}{\textit{Algorithm-specific Arguments}}\\
        \midrule
        \begin{tabular}[c]{@{}c@{}}\textbf{Learning Rate}\\\textbf{Value Network}\end{tabular} & & $10^{-4} \leq x \leq 10^{-1}$ \\
        \greyrule
        \begin{tabular}[c]{@{}c@{}}\textbf{Learning Rate}\\\textbf{Policy Network}\end{tabular} & $3 \cdot 10^{-4} \leq x \leq 3 \cdot 10^{-3}$ & $10^{-4} \leq x \leq 10^{-1}$ \\
        \greyrule
        \begin{tabular}[c]{@{}c@{}}\textbf{Number of}\\\textbf{Training Epochs}\end{tabular} & $x \in \{100, 125, 150, 175\}$ & \\
        \midrule
        \multicolumn{3}{c}{\textit{Koopman-specific Arguments}}\\
        \midrule
        \textbf{Number of Paths} & $x \in \{50, 75, 100, 125, 150, 175, 200 \}$ & $x \in \{50, 75, 100, 125, 150, 175, 200 \}$ \\
        \greyrule
        \textbf{Steps per Path} & $x \in \{75, 100, 125, 150, 175, 200, 225, 250, 275, 300  \}$ &  $x \in \{75, 100, 125, 150, 175, 200, 225, 250, 275, 300 \}$ \\
        \greyrule
        \begin{tabular}[c]{@{}c@{}}\textbf{$\mathcal{O}$(Monomials)}\\\textbf{State Dictionary}\end{tabular} & $1 \leq x \leq 4$ & $1 \leq x \leq 4$ \\
        \greyrule
        \begin{tabular}[c]{@{}c@{}}\textbf{$\mathcal{O}$(Monomials)}\\\textbf{Action Dictionary}\end{tabular} & $1 \leq x \leq 4$ & $1 \leq x \leq 4$ \\
        \bottomrule
\end{tabular}

    }

\end{table}

\noindent All reinforcement learning algorithms either rely, or build on the CleanRL reinforcement learning library~\cite{huang2022cleanrl} to ensure correctness of baselines, and full reproducibility. We utilize the Ray Tune library~\footnote{\href{https://github.com/ray-project/ray/tree/master/python/ray/tune}{github.com/ray-project/ray/tree/master/python/ray/tune}} in conjunction with Optuna~\footnote{\href{https://github.com/optuna/optuna}{github.com/optuna/optuna}} for the hyperparameter optimization. The code to reproduce all experiments is available in the KoopmanRL Python package~\footnote{\href{https://github.com/dynamicslab/KoopmanRL}{github.com/dynamicslab/KoopmanRL}}. The processed dataset of all experiments can be found in the KoopmanRL data repository on HuggingFace~\footnote{\href{https://huggingface.co/datasets/dynamicslab/KoopmanRL}{https://huggingface.co/datasets/dynamicslab/KoopmanRL}}.

\begin{table}[ht]
    \centering
    \resizebox{0.985\textwidth}{!}{

\begin{tabular}{ccccccccc}
    \toprule
    \multirow{2}{*}{{\renewcommand{\arraystretch}{1}\begin{tabular}[c]{@{}c@{}}\textbf{\textbf{Hyperparameter}}\vspace{.0em}\end{tabular}}} & \multicolumn{4}{c}{{\renewcommand{\arraystretch}{1}\begin{tabular}[c]{@{}c@{}}\textbf{\textbf{Soft Koopman Value Iteration}}\vspace{.0em}\end{tabular}}} & \multicolumn{4}{c}{{\renewcommand{\arraystretch}{1}\begin{tabular}[c]{@{}c@{}}\textbf{\textbf{Soft Actor Koopman-Critic}}\vspace{.0em}\end{tabular}}} \\
    \cmidrule(lr){2-5} \cmidrule(lr){6-9}
     & {\renewcommand{\arraystretch}{1}\begin{tabular}[c]{@{}c@{}}\textbf{\textbf{Linear}}\\\textbf{ System}\vspace{.0em}\end{tabular}} & {\renewcommand{\arraystretch}{1}\begin{tabular}[c]{@{}c@{}}\textbf{\textbf{Lorenz}}\\\textbf{1963}\vspace{.0em}\end{tabular}} & {\renewcommand{\arraystretch}{1}\begin{tabular}[c]{@{}c@{}}\textbf{\textbf{Fluid}}\\\textbf{Flow}\vspace{.0em}\end{tabular}} & {\renewcommand{\arraystretch}{1}\begin{tabular}[c]{@{}c@{}}\textbf{\textbf{Stochastic}}\\\textbf{Double Well}\vspace{.0em}\end{tabular}} & {\renewcommand{\arraystretch}{1}\begin{tabular}[c]{@{}c@{}}\textbf{\textbf{Linear}}\\\textbf{System}\vspace{.0em}\end{tabular}} & {\renewcommand{\arraystretch}{1}\begin{tabular}[c]{@{}c@{}}\textbf{\textbf{Lorenz}}\\\textbf{1963}\vspace{.0em}\end{tabular}} & {\renewcommand{\arraystretch}{1}\begin{tabular}[c]{@{}c@{}}\textbf{\textbf{Fluid}}\\\textbf{ Flow}\vspace{.0em}\end{tabular}} & {\renewcommand{\arraystretch}{1}\begin{tabular}[c]{@{}c@{}}\textbf{\textbf{Stochastic}}\\\textbf{Double Well}\vspace{.0em}\end{tabular}}\\
    \midrule
    \multicolumn{9}{c}{\textit{Algorithm-specific Arguments}}\\
    \midrule
    \begin{tabular}[c]{@{}c@{}}\textbf{Total}\\\textbf{Timesteps}\end{tabular} & \multicolumn{1}{|c}{50,000} & 50,000 & 50,000 & 50,000 & \multicolumn{1}{|c}{50,000} & 50,000 & 50,000 & 50,000 \\
    \greyrule
    \textbf{Buffer Size} & \multicolumn{1}{|c}{} & & & & \multicolumn{1}{|c}{1,000,000} & 1,000,000 & 1,000,000 & 1,000,000 \\
    \greyrule
    \textbf{Gamma} & \multicolumn{1}{|c}{0.99} & 0.99 & 0.99 & 0.99 & \multicolumn{1}{|c}{0.99} & 0.99 & 0.99 & 0.99 \\
    \greyrule
    \textbf{Tau} & \multicolumn{1}{|c}{} & & & & \multicolumn{1}{|c}{0.005} & 0.005 & 0.005 & 0.005 \\
    \greyrule
    \textbf{Batch Size} & \multicolumn{1}{|c}{16384} & 16384 & 16384 & 16384 & \multicolumn{1}{|c}{256} & 256 & 256 & 256 \\
    \greyrule
    \textbf{Learning Start} & \multicolumn{1}{|c}{} & & & & \multicolumn{1}{|c}{5,000} & 5,000 & 5,000 & 5,000 \\
    \greyrule
    \begin{tabular}[c]{@{}c@{}}\textbf{Policy}\\\textbf{Learning Rate}\end{tabular} & \multicolumn{1}{|c}{} & & & & \multicolumn{1}{|c}{0.0003} & 0.0003 & 0.0003 & 0.0003 \\
    \greyrule
    \begin{tabular}[c]{@{}c@{}}\textbf{Value Network}\\\textbf{Learning Rate}\end{tabular} & \multicolumn{1}{|c}{} & & & & \multicolumn{1}{|c}{\textbf{\textcolor{sac_q_col}{0.00047}}} & \textbf{\textcolor{sac_q_col}{0.05157}} & \textbf{\textcolor{sac_q_col}{0.0094}} & \textbf{\textcolor{sac_q_col}{0.00033}} \\
    \greyrule
    \begin{tabular}[c]{@{}c@{}}\textbf{Policy Network}\\\textbf{Learning Rate}\end{tabular} & \multicolumn{1}{|c}{\textbf{\textcolor{sac_q_col}{0.00109}}} & \textbf{\textcolor{sac_q_col}{0.00051}} & \textbf{\textcolor{sac_q_col}{0.00032}} & \textbf{\textcolor{sac_q_col}{0.00166}} & \multicolumn{1}{|c}{\textbf{\textcolor{sac_q_col}{0.0018}}} & \textbf{\textcolor{sac_q_col}{0.0236}} & \textbf{\textcolor{sac_q_col}{0.0018}} & \textbf{\textcolor{sac_q_col}{0.0004}} \\
    \greyrule
    \textbf{Policy Frequency} & \multicolumn{1}{|c}{} & & & & \multicolumn{1}{|c}{2} & 2 & 2 & 2 \\
    \greyrule
    \begin{tabular}[c]{@{}c@{}}\textbf{Target Network}\\\textbf{Frequency}\end{tabular} & \multicolumn{1}{|c}{} & & & & \multicolumn{1}{|c}{1} & 1 & 1 & 1 \\
    \greyrule
    \textbf{Noise Clip} & \multicolumn{1}{|c}{} & & & & \multicolumn{1}{|c}{0.5} & 0.5 & 0.5 & 0.5 \\
    \greyrule
    \textbf{Alpha} & \multicolumn{1}{|c}{1.0} & 1.0 & 1.0 & 1.0 & \multicolumn{1}{|c}{0.2} & 0.2 & 0.2 & 0.2 \\
    \greyrule
    \begin{tabular}[c]{@{}c@{}}\textbf{Alpha}\\\textbf{Learning Rate}\end{tabular} & \multicolumn{1}{|c}{} & & & & \multicolumn{1}{|c}{0.001} & 0.001 & 0.001 & 0.001 \\
    \greyrule
    \begin{tabular}[c]{@{}c@{}}\textbf{Number of}\\\textbf{Actions}\end{tabular} & \multicolumn{1}{|c}{101} & 101 & 101 & 101 & \multicolumn{1}{|c}{} & & & \\
    \greyrule
    \begin{tabular}[c]{@{}c@{}}\textbf{Number of}\\\textbf{Training Epochs}\end{tabular} & \multicolumn{1}{|c}{\textbf{\textcolor{sac_q_col}{125}}} & \textbf{\textcolor{sac_q_col}{125}} & \textbf{\textcolor{sac_q_col}{125}} & \textbf{\textcolor{sac_q_col}{175}} & \multicolumn{1}{|c}{} & & & \\
    \greyrule
    \textbf{Batch Scale} & \multicolumn{1}{|c}{1} & 1 & 1 & 1 & \multicolumn{1}{|c}{} & & & \\
    \greyrule
    \textbf{Autotune} & \multicolumn{1}{|c}{} & & & & \multicolumn{1}{|c}{True} & True & True & True \\
    \greyrule
    \textbf{Regressor} & \multicolumn{1}{|c}{OLS} & OLS & OLS & OLS & \multicolumn{1}{|c}{OLS} & OLS & OLS & OLS \\
    \midrule
    \multicolumn{9}{c}{\textit{Koopman-specific Arguments}}\\
    \midrule
    \textbf{Number of Paths} & \multicolumn{1}{|c}{\textbf{\textcolor{sac_q_col}{75}}} & \textbf{\textcolor{sac_q_col}{150}} & \textbf{\textcolor{sac_q_col}{200}} & \textbf{\textcolor{sac_q_col}{175}} & \multicolumn{1}{|c}{\textbf{\textcolor{sac_q_col}{150}}} & \textbf{\textcolor{sac_q_col}{200}} & \textbf{\textcolor{sac_q_col}{50}} & \textbf{\textcolor{sac_q_col}{150}} \\
    \greyrule
    \textbf{Steps per Path} & \multicolumn{1}{|c}{\textbf{\textcolor{sac_q_col}{250}}} & \textbf{\textcolor{sac_q_col}{250}} & \textbf{\textcolor{sac_q_col}{225}} & \textbf{\textcolor{sac_q_col}{100}} & \multicolumn{1}{|c}{\textbf{\textcolor{sac_q_col}{175}}} & \textbf{\textcolor{sac_q_col}{150}} & \textbf{\textcolor{sac_q_col}{175}} & \textbf{\textcolor{sac_q_col}{300}} \\
    \greyrule
    \begin{tabular}[c]{@{}c@{}}\textbf{$\mathcal{O}$(Monomials)}\\\textbf{State Dictionary}\end{tabular} & \multicolumn{1}{|c}{\textbf{\textcolor{sac_q_col}{2}}} & \textbf{\textcolor{sac_q_col}{3}} & \textbf{\textcolor{sac_q_col}{4}} & \textbf{\textcolor{sac_q_col}{2}} & \multicolumn{1}{|c}{\textbf{\textcolor{sac_q_col}{2}}} & \textbf{\textcolor{sac_q_col}{2}} & \textbf{\textcolor{sac_q_col}{3}} & \textbf{\textcolor{sac_q_col}{4}} \\
    \greyrule
    \begin{tabular}[c]{@{}c@{}}\textbf{$\mathcal{O}$(Monomials)}\\\textbf{Action Dictionary}\end{tabular} & \multicolumn{1}{|c}{\textbf{\textcolor{sac_q_col}{3}}} & \textbf{\textcolor{sac_q_col}{1}} & \textbf{\textcolor{sac_q_col}{2}} & \textbf{\textcolor{sac_q_col}{4}} & \multicolumn{1}{|c}{\textbf{\textcolor{sac_q_col}{3}}} & \textbf{\textcolor{sac_q_col}{1}} & \textbf{\textcolor{sac_q_col}{3}} & \textbf{\textcolor{sac_q_col}{4}} \\
    \bottomrule
\end{tabular}

    }
    \caption{
        Optimal hyperparameter configurations found through the computational constraints placed on the workflow. Optimized parameters are highlighted in pink. Templated arguments, which were not included in the tuning are provided for context. Each environment was optimized for individually. \textit{OLS} as regression algorithm refers to the \textit{Ordinary Least Squares} regression algorithm. For the full default parameter template, we refer the reader to the respective single-file implementations in the Python package.
    }
    \label{tab:hyperparameter_configuration}
\end{table}

\section{Results}
\label{Results}

We begin by examining the performance of the Koopman-assisted reinforcement learning algorithms on the four benchmark environments. Subsequently we will perform a series of ablation studies, before concluding with an exploration of the interpretability inherent to the Koopman-assisted reinforcement learning approach. The linear system environment is excluded from the episodic return, and application of control plot due to all control approaches being statistically indistinguishable from each other. The results of the linear system are available in appendix~\ref{appendix:episodic_returns_linear_system}. 

\subsection{Algorithm Performance on Benchmark Problems}

\begin{figure}[t]
    \centering
    \includegraphics[width=.85\textwidth]{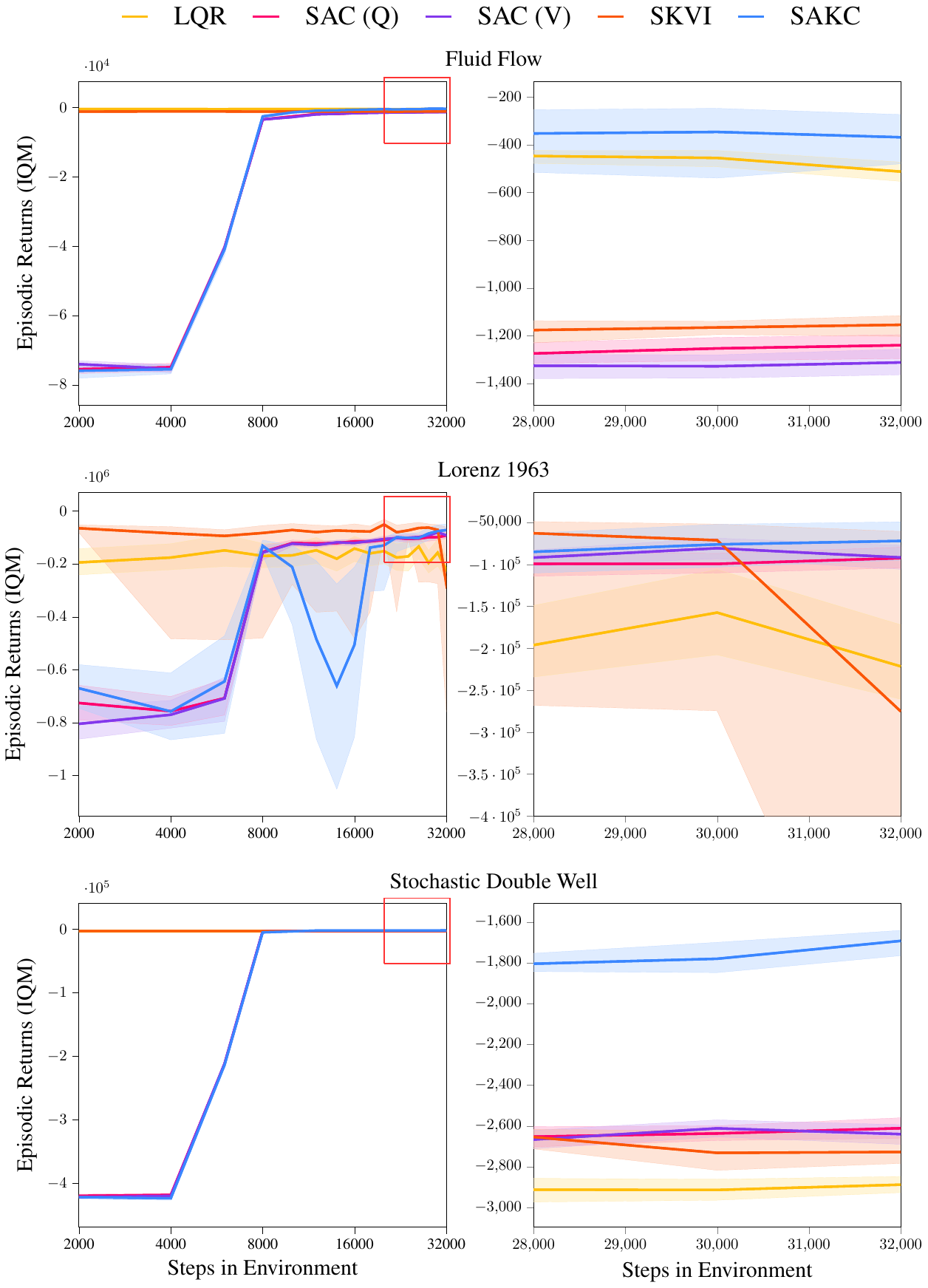}
    \caption{
        Algorithm performance as measured by the inter-quartile means of the episodic returns with the $95\% $ confidence band reported around each measurement. \textbf{(1st Column)} Performance over the entire timespan of the control experiment applied to the environment for each of the control candidates. Highlighted with red squares are cutouts. \textbf{(2nd Column)} Cutout of the performance in the last $4,000$ steps to discern the final performance of the algorithms.
    }
    \label{fig:iqm-returns}
\end{figure}

\begin{figure}[t]
    \centering
    \includegraphics[width=\textwidth]{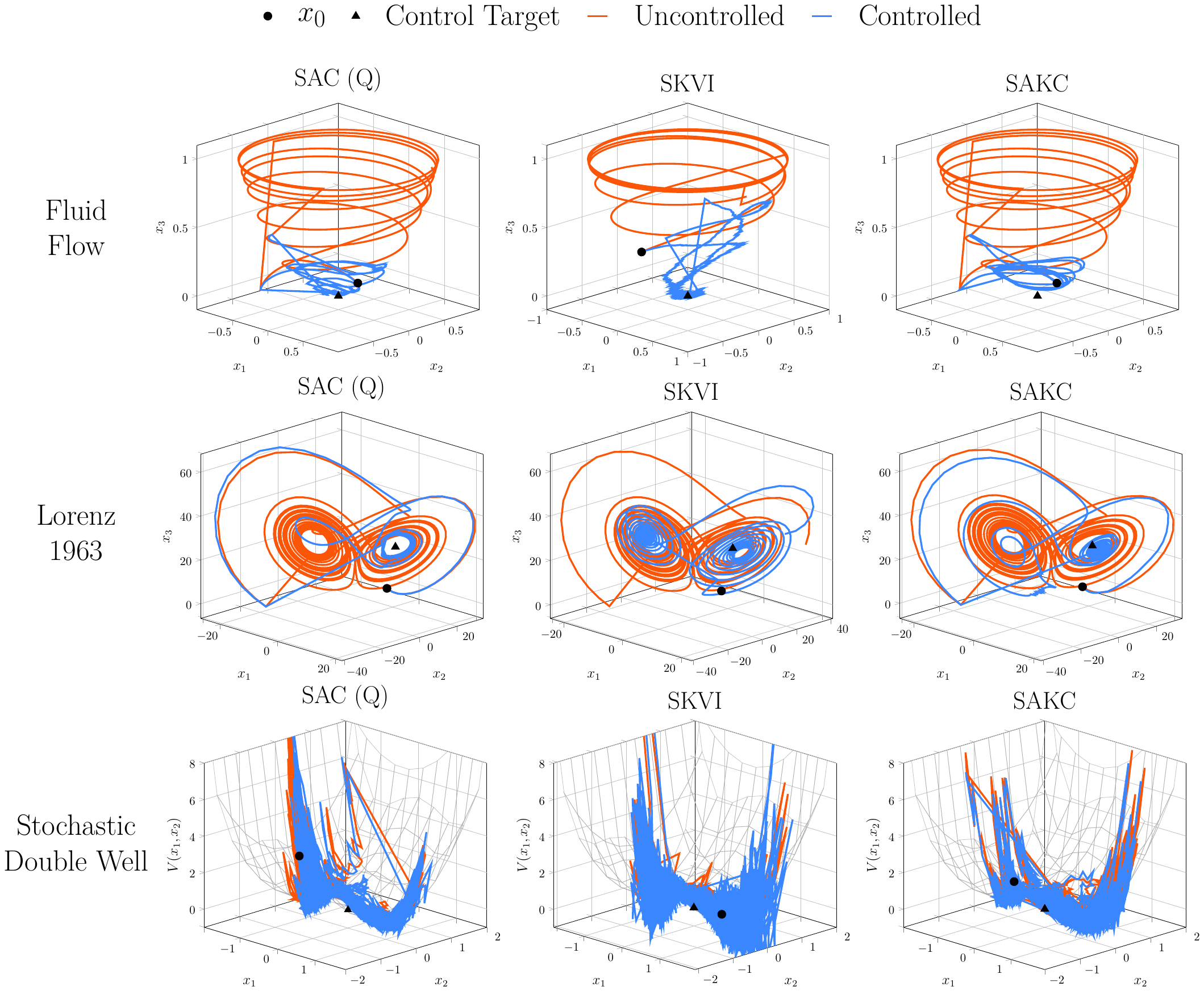}
    \caption{
        Evolution of the control being applied over the course of time with the controlled evolution (blue), and the uncontrolled evolution of the dynamical system (orange). The target control for each environment is plotted as the black triangle. The vector field of the potential surface is plotted as additional context for the double-well system.
    }
    \label{fig:control_applied}
\end{figure}

The performance of Soft Actor Koopman-Critic is evaluated on the four environments and compared against the classical control baseline, as well as the reference SAC (Q) and SAC (V) algorithms. 
Comparisons are provided in Fig.~\ref{fig:iqm-returns}, and the control roll-outs are provided in Fig.~\ref{fig:control_applied}. 
We see that on all environments, SAKC needs $\approx 8000$ steps in the environment to properly initialize the Koopman tensor and calibrate its Koopman critic to match the SOTA performance of LQR, the soft Koopman value iteration, and SAC (V).

On the \textit{Fluid Flow} environment, this training dynamic is matched, with SAKC and SAC (V) requiring $\approx 8000$ environment steps to explore the environment. Importantly, SAKC beats SOTA in this more difficult environment. To better inspect the performance differences between the tested algorithms in this more challenging environment, we provide a zoomed-in excerpt covering $28,000$ to $32,0000$ steps in Fig.~\ref{fig:iqm-returns}, column b. The performance of LQR stays stably at a high level, with it notably outperforming the actor-critic variants in the low resource setting with $\leq 8,000$ steps in the environment. Similar low-interaction outperformance applies to the Soft Koopman Value Iteration which shines in the same interval as LQR, but is unable to match the performance of LQR on this particular environment. The actor-critic based reinforcement learning algorithms begin to more closely match the performance with more steps in the environment. Yet, the final performance of the Soft Actor Koopman-Critic is beyond the performance of the Soft Actor-Critic variants with LQR being at the lower end of the Soft Actor Koopman-Critic's confidence interval.

On the chaotic \textit{Lorenz 1963}, we see again that the actor critic variants need around $\approx 8,000$ steps to learn enough about the environment to be able to control it well. Most notably, on this chaotic environment the Soft Koopman Value Iteration performs best for the first $\approx 30,000$ steps in the environment, seeing that its Koopman representation calibrated exceedingly well, but then beyond $30,000$ steps sees a steep fall in its performance which we attribute to the offline Koopman construction procedure, which in the case of a chaotic system would require online reconstruction once the confidence in the learned Koopman representation falls below a threshold value. We leave this avenue of improvement to future work. After the initial learning steps, all 3 actor-critic variants perform within the statistical confidence intervals of each other, and outperform the value iteration, as well as LQR. While the Soft Actor Koopman-Critic outperforms the state-of-the-art set by the soft actor-critic algorithms, its confidence interal overlaps with the ones by the actor-critics and as such the performance essentially matches the state-of-the-art.

On the \textit{Stochastic Double Well}, we see the typical $\approx 8,000$ steps in the environment learning period of the actor-critic variants, beyond which they begin to match, and then outperform the LQR and Soft Koopman Value Iteration. As we can see in the cutout, over time 3 distinct performance groups develop. The Soft Actor Koopman-Critic outperforming all other approaches, a second group of the soft actor-critic variants, and the Soft Koopman Value Iteration perform slightly worse, with the LQR only lagging slightly behind in performance.

In general, we can conclude from all 3 environments, that LQR and the Soft Koopman Value Iteration always outperform the actor critic variants in the low steps in environment limit. Of the 3 actor-critic variants, the Soft Actor Koopman-Critic provides the best performance on the tested environments, but the generalization of this finding to more general reinforcement learning benchmark suites like Atari~\cite{bellemare2013arcade}, and Mujoco~\cite{todorov2012mujoco} to substantiate that claim is left for future work.

When examining the prolonged control performance of the trained algorithms on the Fluid Flow environment, the Lorenz system, and stochastic double well, we constrain ourselves to the best-performing reinforcement learning algorithm according to the episodic returns in the form of the Q-function formulation-based soft actor-critic algorithm, comparing its performing to the two Koopman-based variants in figure~\ref{fig:control_applied}. On the fluid flow, the two Koopman-variants both provide better performance than the SAC-variant with their controlled trajectories much closer to the control target. Especially the soft Koopman value iteration shines in this instance with a highly compact trajectory close the control target, the authors surmise this being the result of the learned Koopman operator's modes being well-matched to the fluid flow's dominant modes. Of the three variants the soft actor-critic variant provides the worst performance with a noisier trajectory with high variance, and multiple trajectory divergences from the control target within the examined time frame. On the Lorenz system, soft actor-critic and the soft actor Koopman-critic both provide much tigher control trajectories, than the soft Koopman value iteration which struggles to lock in on the control target's lobe of the Lorenz system. Both the soft actor-critic, and the soft actor Koopman-critic suffer much less from this behaviour. The stochastic double well poses a significant challenge for all three approaches. Their control trajectories are noisy, and exhibit limited directional control to converge onto the control target with a minor performance advantange of the soft actor Koopman-critic over the other two algorithms improving upon the uncontrolled baseline.

\subsection{Ablation Analyses}
\label{subsec:ablation-analysis}

With the already ample hyperparameter configurations of the value iteration approach, and the soft actor-critic algorithms, the Koopman guidance introduces an extra set of hyperparameters who have to be configured appropriately, and whose interplay with the other hyperparameters warrants quantitative exploration. The vast combinatoric space opened up with this number of hyperparameters is on display in the tuned hyperparameter configurations, see table~\ref{tab:hyperparameter_configuration}. We begin by examining the impact of the interplay between the number of actions, and the number of training epochs afforded to the Soft Koopman Value Iteration. Subsequently, the interplay between the value network learning rate, and the policy network learning is examined for the Soft Actor Koopman-Critic. The presented ablation studies only present a glimpse into the combinatorial space of potential hyperparameter choices of KARL algorithms, the full exploration of which is beyond the scope of the present paper.

\subsubsection{Soft Koopman Value Iteration}

\begin{figure}[t]
    \centering
    \includegraphics[width=\textwidth]{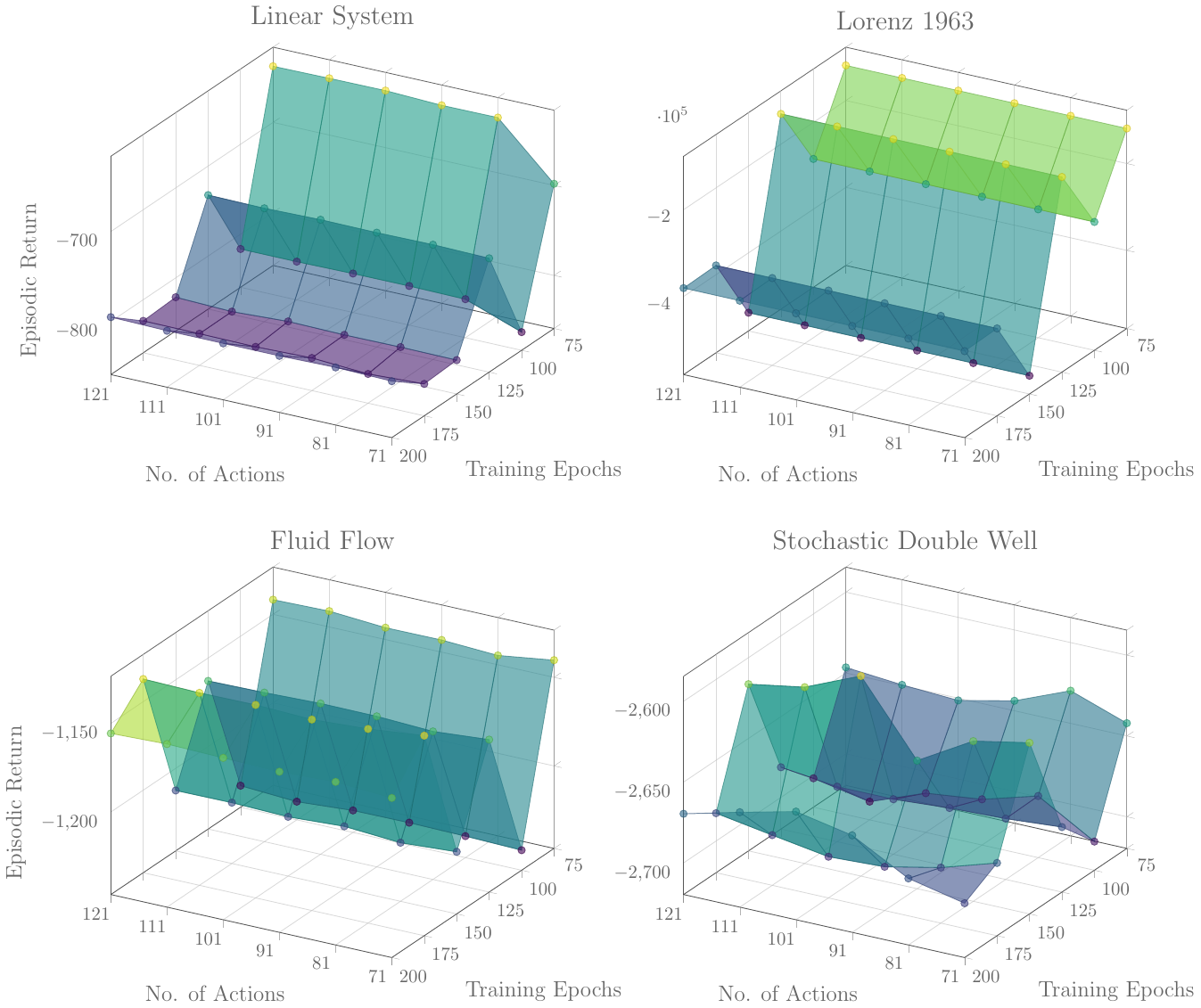}
    \caption{
        Ablation of the influence of the \textit{number of actions} and the number of \textit{training epochs} afforded to the construction of the Koopman operator on the performance of the \textit{Soft Koopman Value Iteration} performed for all four environments.
    }
    \label{fig:skvi-ablations}
\end{figure}

Examining the interplay of the number of actions, and the number of training epochs afforded a number of interesting insights arise. First and foremost it becomes apparent when inspecting the data visualized as a 3D-surface plots in Fig.~\ref{fig:skvi-ablations} that the number of actions afforded to the algorithm has no tangible impact on the algorithm's performance. In the case of the linear system, lorenz 1963, and the fluid flow environment the episodic return stays constant across the number of actions, and only the stochastic double well shows a sensitivity to the number of actions with $101$ actions performing the best. The number of trainings epochs displays a much larger influence with less training epochs tending to result in better long-term performance. The authors surmise that an overly large number of training epochs results in the operator overfitting to the dynamics, and subsequently being overfit to the dynamics sampled by the random agent. This hypothesis would be in line with the wider machine learning for dynamical systems literature in which the issue of overfitting to (early) dynamics is an ever present issue. In the case of the stochastic double well environment, no clear trend is discernible with the local optima being at $150$ training epochs. Similar constraints apply to the fluid flow with both $75$, and $200$ training epochs resulting in good performance but all grid points being within $\approx 8 \%$ performance of each other i.e. there being little to no difference in terms of performance.

\subsubsection{Soft Koopman Actor-Critic}

Just like for the Soft Koopman Value Iteration, the choice of hyperparameters governing the Koopman operator, are a set of major hyperparameters governing the performance of the algorithm. These choice have been explored implicitly in the hyperparameter optimization, see table~\ref{tab:hyperparameter_configuration}. In the case of the Soft Actor Koopman-Critic we specifically focus on the interplay of the value network learning rate, and the policy network learning rate. The reason for this being that the default learning rates were adapted from the Soft Actor-Critic learning rates previously, yet the introduction of the Koopman operator component warrants revisiting.

Across all four environments it is apparent that both, the value network learning rate as well as the policy network learning rate have a direct influence on the performance of the algorithm. In the case of the the fluid flow, as well as the stochastic double well environment a large value network learning rate in conjunction with a medium policy network learning rate in the range $[ 0.01, 0.0005 ]$ produced the best performance. Lower value network see an especially large fall-off in performance for the fluid flow, whereas performance stays stable in the case of the stochastic double well. A similar decline in performance with a smaller value network learning rate is visible in the case of the linear system, which sees a plateau of high performance at low value network learning rates irrespective of the policy network learning rate. In the case of the Lorenz 1963 environment it is not possible to draw strict conclusions of the interplay between the two learning rates at this point, with the used grid sampling space. Both extrema of the grid sampling space do produce good performance with the two absolute extrema of the sampling space i.e. both learning rates being large, and both learning rates being small seemingly producing good performance. This behavior might be attributable to the highly chaotic nature of the Lorenz environment. To provide conclusive insight into this interplay on the Lorenz environment, a much larger sampling space would have to be utilized, equally to rule out a "fractal" nature of the episodic return landscape, one would have to sample the space the between the grid sampling point to accuractely capture those finer structures of the space. This exploration is beyond the computational resources of ours.

\begin{figure}[t]
    \centering
    \includegraphics[width=\textwidth]{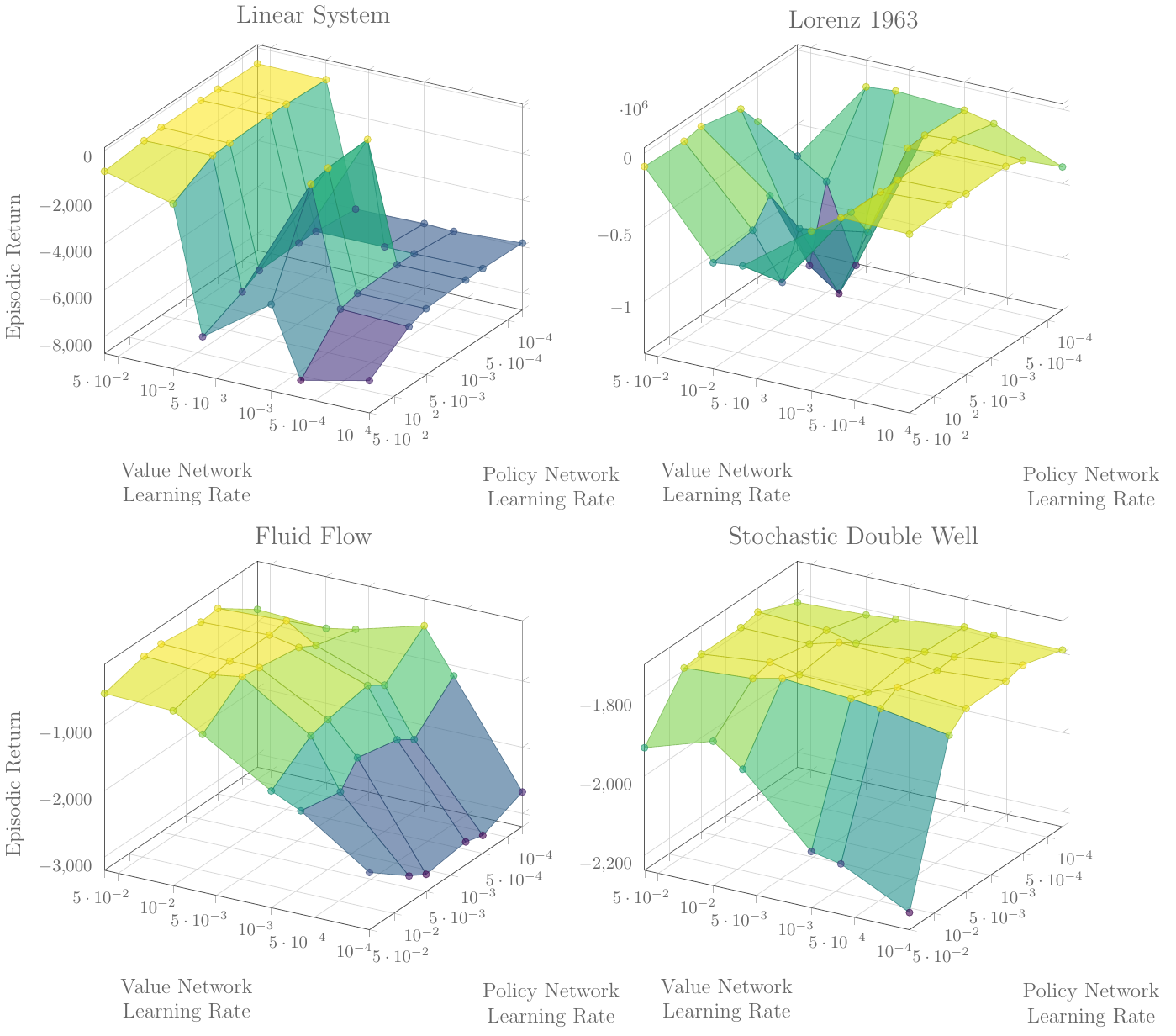}
    \caption{
        Ablations of the influence of the \textit{Policy Learning Rate} and the \textit{Value Network Learning Rate} on the performance of the \textit{Soft Actor Koopman-Critic} performed for each of the four environment individually
    }
    \label{fig:sakc-ablations}
\end{figure}

\noindent In conclusion, outside of the Lorenz system, we can conclude that one can expect the best performance from a large value network learning rate of $0.05$ or potentially even larger, and a policy network learning rate in the range of $[0.01, 0.0005]$ with the Lorenz system producing a better performance when one picks a policy network learning rate at the upper end of that range.

\subsection{Interpretability of Koopman Reinforcement Learning}

A core feature of Koopman-assisted reinforcement learning is that the construction approach taken to the Koopman operator introduces a level of inherent interpretability which permits the user to not only introspect the algorithm's behaviour, but go as far as influencing the RL policy's behavior. This is not possible with standard opaque optimal value function modeling methods. To demonstrate this, we're inspecting the \textit{Soft Koopman Value Iteration} on the four environments, where leverage the candidate-function space comprised of identifiable component terms and, in turn, the array of coefficients for those terms corresponding to the optimal value function.

\begin{table}[t]
    \centering
    \caption{
        Polynomials extracted from each of the value functions of the best-performing \textit{Soft Koopman Value Iterations} on the four environments. 
    }
    \label{tab:value-func-interpretability}
    \resizebox{\textwidth}{!}{

\begin{tabular}{cc}
        \toprule
        {\renewcommand{\arraystretch}{1}\begin{tabular}[c]{@{}c@{}}\textbf{\textbf{Environment}}\vspace{.0em}\end{tabular}} & {\renewcommand{\arraystretch}{1}\begin{tabular}[c]{@{}c@{}}\textbf{\textbf{Soft Koopman Value Iteration}}\vspace{.0em}\end{tabular}} \\
        \midrule
        $\begin{array}{c}
             \text{Linear System} \\
              \text{(3D)}
        \end{array}$ & $\begin{array} {c}
            1.19 x^{2} +1.01 y^{2} +1.07 z^{2} +0.07 xy +0.23 xz +0.05 yz \\
            -0.01 x +0.01 y -0.02 z -185.32
        \end{array}$ \\
        \greyrule $\begin{array}{c}
             \text{Fluid Flow} \\
             \text{(3D)}
        \end{array}$ & $\begin{array}{c}
             16.69 x^{4} +0.27 y^{4} +0.08 z^{4} -4.88 x^{3}y -11.82 x^{3}z +15.06 xy^{3} \\
             +1.08 y^{3}z -6.98 xz^{3} +0.36 yz^{3} -0.43 x^{2}y^{2} -1.44 x^{2}yz +0.02 xy^{2}z \\
             +0.02 x^{2}z^{2} +3.28 xyz^{2} +0.25 y^{2}z^{2} +0.25 x^{3} -0.08 y^{3} +7.88 z^{3} \\
             -0.31 x^{2}y +46.61 xy^{2} -5.98 x^{2}z +0.42 xyz -0.34 y^{2}z +29.87 xz^{2} \\
             -1.04 yz^{2} +132.62 x^{2} +99.59 y^{2} +37.92 z^{2} -24.51 xy +0.17 xz \\
             +0.5 yz -0.19 x +0.23 y -1.47 z -260.17
        \end{array}$ \\
        \greyrule
        $\begin{array}{c}
            \text{Lorenz 1963} \\
            \text{(3D)}
        \end{array}$ & $\begin{array}{c}
            2.88 x^{3} +0.33 y^{3} +0.19 z^{3} +0.79 x^{2}y -0.39 xy^{2} +0.36 x^{2}z \\
            +0.22 xyz +0.99 y^{2}z -0.38 xz^{2} -0.51 yz^{2} -7.65 x^{2} +12.74 y^{2} \\
            -12.81 z^{2} -13.2 xy +13.46 xz +18.78 yz -1024.40 x -1225.20 y \\
            +486.33 z - 1431.40 
        \end{array}$ \\
        \greyrule
        $\begin{array}{c}
            \text{Stochastic Double Well}  \\
            \text{(2D)}
        \end{array}$ & $\begin{array}{c}
            8.41 x^{2} +24.49 y^{2} +1.23 xy -0.30 x +3.20 y -115.18
        \end{array}$ \\
        \bottomrule
\end{tabular}

    }
\end{table}

By revealing the relative contributions of each term, in contrast to non-interpretable models, KARL supports a more informed approach to tailoring value functions. To this end one is able to filter out individual terms based on thresholds, or based on particular insight into the environment itself. For example, our extensive interaction with fluid flow simulations indicated that forcing the $y$ spatial coordinate had more impact than other components. This enables an explicit trade-off between computational complexity of the value function and wall-clock time optimization of computation for the user. Thus, expert knowledge, guided by interpretability, enables the user to identify the importance of simpler, and potentially more generalizable, value functions.  Several learned value functions are shown in Tab.~\ref{tab:value-func-interpretability}.

This example case study shows KARL's ability to enable more interpretability in RL for guiding efforts to simplify and improve model output and to support adaptability to real-world environments with imperfect simulators or unsatisfactory reward functions. By allowing experts to limit dependence on less important terms, KARL promotes value functions focused on more crucial outcome drivers across diverse scenarios, facilitating prospects for more generalizable policies.

\section{Limitations and Future Work}
\label{Future_Work}

While Koopman-assisted reinforcement learning is promising, it faces in its current form a number of limitations, which are subject to future work. We first note in subsection~\ref{dictionary_dependence} KARL's crucial dependence on the specification of the dictionary space, and how this aspect of KARL can be extended by considering dictionary spaces paired with regularization methods. Another inherent constraint of KARL is the current \textit{offline}, sampling-reliant construction of the Koopman tensor, and thus the action-dependent Koopman operator in the feature space. In subsection~\ref{online_learning} we outline how this could potentially be addressed with an \textit{online}, recursive algorithm. Next, KARL is currently limited to discrete-time environments/dynamical systems, and its potential extension to continuous-time systems is discussed in subsection~\ref{koopman_generator}. Finally, SKVI currently relies on a discretization of the action space for which we briefly discuss the limitations as well as an approach we are developing to allow for the more robust continuous action setting. In the following we will delve in detail into these limitations, and their potential for future work.

\subsection{Dictionary Dependence and Discovery}\label{dictionary_dependence}
The Koopman approach as outlined above, and further explored in the performance results in section~\ref{Results}, has a crucial dependence on the correct choice of dictionary space. This dictionary space must both characterize the value function and the dynamic evolution of the value function. Specifically, we require that different linear combinations of the elements of the vector $\phi(x)$ can approximate $V(x)$ and $V(x')$ well. There are, however, a number of methods that can be incorporated to help relax the specification of a correct dictionary space. First, to address the problem of finding the Koopman tensor, we recall the regression problem \eqref{eq:fit-koopman-tensor}. As this is a standard regression problem, it lends itself well to regularization, such as ridge regression, LASSO regression, or other custom regularizers. Second, to address the problem of finding the representation of the value function at the current state, we can again utilize a large dictionary space while leveraging the aforementioned regression sparsification methods. In particular, the regression problems we face for the problem of fitting the contemporaneous value function in our KARL algorithms are \eqref{eq:be} and \eqref{eq:soft_value_loss}. Again each of these can be treated as a traditional least squares regression problem and admit the use of traditional regularization methods, such as a sparsification.

\subsection{Online Learning of the Koopman Tensor}
\label{online_learning}

One concern around the construction of the Koopman tensor is its reliance on a sufficiently rich set of static training data. This is problematic in online settings. To address this we see great potential in building on the recursive EDMD (rEDMD) algorithm developed by Sinha et al.~\cite{sinha2019online}. The exploration of this direction is left to future work.

\subsection{Continuous-Time Settings: Koopman Generator Approach}
\label{koopman_generator}

Here we recall a useful object for continuous time stochastic processes, the generator of the Koopman operator $\mathcal{L}$. This operator is the time derivative of the Koopman operator evaluated at time $t=0$ for the time homogenous case. The analogous object in discrete time is the transition matrix of $\tilde X$. First, we represent the dynamics of the stochastic process as a stochastic differential equation (SDE) which we will assume is continuous itself, i.e., has no jumps:
\begin{subequations}
\begin{align}
    \text{d}X_t &= \mu_x(X_t, U_t)\text{d}t + \sigma_x(X_t,U_t)\text{d}W_t\\
    \text{d}U_t &= \mu_u(X_t)\text{d}t + \sigma_u(X_t)\text{d}W_t .
\end{align}
\end{subequations}
Here, $\mu_x: \mathbb{X} \times \mathbb{U} \to \mathbb{R}^n$ and $\mu_u: \mathbb{X} \to \mathbb{R}^m$ are the drift terms and $\sigma_x: \mathbb{X} \times \mathbb{U} \to \mathbb{R}^{n\times d}$ and $\sigma_u: \mathbb{X}  \to \mathbb{R}^{m\times d}$ are the diffusion terms, for the state and control, respectively. Finally, $W_t$ is a $d$-dimensional Wiener process. Given a twice continuously differentiable function $f$, it can be shown using It\^{o}'s lemma that the infinitesimal generator of the Koopman operator is characterized by
\begin{equation}
    \mathcal{L}^uf = \mu_x\cdot\nabla_{x}f + \frac{1}{2}a_x:\nabla^2_{x}f
\end{equation}
where $a_x = {\sigma_x} {\sigma_x}^\top$, $\nabla^2_x$ denotes the Hessian, and $:$ denotes the double dot product. In this setting the function $k(t, x) = \mathcal{K}_tf(x)$ satisfies the second-order partial differential equation $\frac{\partial k}{\partial t} = \mathcal{L}k$, which is called the Kolmogorov backward equation. See~\cite{klus2020data} for more details. Adapting tools from~\cite{klus2020data} we can find the Koopman generator tensor by solving the following least squares problem
\begin{align}
    \min_{L\in \mathbb{R}^{d_x\times (d_x d_u)}} \sum_{i=1}^N \left\|  L(\psi(u_i)\otimes \phi(x_i)) - \widehat{\mathcal{L}}
    ^{u_i}\phi(x_i)  \right\|^2
\end{align}
where 
\begin{align}
    \widehat{\mathcal{L}}^{u_i}\phi = \hat\mu_x\cdot\nabla_{x}\phi + \frac{1}{2}\hat a_x:\nabla^2_{x}\phi
\end{align}
with
\begin{subequations}
\begin{align}
    \hat \mu_x &:= \frac{\Delta x_i}{\Delta_t} = \frac{x_{i+1}-x_i}{t_{i+1}-t_i} 
    \\
    \hat a_x &:= \frac{(\Delta x_i)(\Delta x_i)^{\top}}{\Delta_t} = \frac{(x_{i+1}-x_i)(x_{i+1}-x_i)^{\top}}{(t_{i+1}-t_i)}
\end{align}
\end{subequations}
and $L\in \mathbb{R}^{d_x\times (d_x d_u)}$ is the flattened Koopman generator tensor where $d_x$ and $d_u$ are the dimensions of the state and action feature map dimensions, respectively.
As with the discrete-time Koopman tensor, once we compute $L$, we can extract a Koopman generator operator for any $u \in \mathcal{U}$ by reshaping $L$ to the 3D tensor $T_L$. Then the $d_x\times d_x$ finite dimensional Koopman generator operator approximation is again $L^{u}$  for any $u \in \mathcal{U}$ as seen in the summation in Fig.~\ref{fig:koopman_tensor}, i.e. $L^{u}[i,j] = \sum_{z = 1}^d T_L(i, j, z) \psi(u)[z]$.

\subsection{Continuous Actions in Soft Koopman Value Iteration}
\label{continuous_action_SKVI}

In our current implementation of SKVI, we discretize the action space and explicitly calculate the mean of the value function evaluated at the future state. Discretizing the action space facilitates a simpler calculation of this mean, as the resulting action distribution that we average over is the softmax distribution. In continuous action space, calculating the mean of the value function evaluated at the future state would require the calculation of a complicated integral for each step of the learning algorithm, due to the softmax policy not necessarily being able to be expressed as a well-known distribution, such as the normal distribution.
The discretization approach we followed has the drawback of correctly "coarse-graining" the action space which requires additional tuning. In order to circumvent this, we have begun work on a continuous action version of the algorithm which is inspired by the actor step in the SAC approach. In particular, we take the additional step of finding the nearest Gaussian distribution to the soft-policy in the KL divergence sense. The results of this implementation are the subject of future work.

\section{Conclusion and Discussion}
\label{Conclusion}
In this work we developed, and introduced two novel Koopman assisted reinforcement learning algorithms by integrating Koopman operator methods with existing maximum entropy RL algorithms. 
By leveraging the Koopman operator, KARL achieves state-of-the-art performance on several challenging benchmark environments. Among its two variants, Soft Actor Koopman-Critic manages to consistently outperform soft actor-critic variants on the benchmark variants lending further credence to the utility of the Koopman operator in improving reinforcement learning approaches. The Soft Koopman Value Iteration meanwhile matches the performance of state-of-the-art algorithms while introducing an element of interpretability which opens significant avenues for future work. KARL's success in these environments demonstrates its flexibility across varying systems: fluid flow shows its prowess in non-linear systems; Lorenz proves KARL's ability to control chaotic systems; and the double well shows its ability to learn in stochastic environments. Because the Koopman operator provides an embedding where strongly nonlinear dynamics become linear, there are several opportunities to leverage more advanced Koopman embeddings~\cite{Brunton2017natcomm,Brunton2022siamreview,colbrook2023mpedmd,colbrook2024rigorous} for improved KARL performance in the future. KARL overcomes many of the limitations of traditional RL algorithms by making them more ``input-output'' interpretable, allowing for a deeper understanding of the learned policies and their underlying dynamics.

The future of KARL lies in its continuous evolution and adaptation to more complex and realistic settings. Addressing these challenges and exploring these directions will allow the Koopman operator to aid in the development of robust, interpretable, and efficient future algorithms and become an integral component of future reinforcement learning approaches.

Prospects for further development and application of KARL are both numerous and promising. For example, integration of KARL with modern online learning techniques~\cite{sinha2019online} could support real-time applications, especially in combination with techniques to improve the efficiency of the algorithm, such as knowledge gradients~\cite{frazier2008knowledge}. A comprehensive theoretical analysis of KARL algorithms, including convergence properties and sample complexity bounds, would provide valuable insights into their behavior, and would aid in providing more intuition and guarantees for safety-critical applications. Regularization, such as sparsity promoting penalties, may facilitate the use of larger dictionary spaces in unfamiliar complex systems.  This may also help to determine value function dynamics and further interpret the main driving features (observables) of the optimal value function. Developing visualization techniques to better interpret the learned Koopman tensor and Koopman-dependent operators may also facilitate broader adoption in domains where interpretability is critical, such as healthcare, economics, and autonomous driving.

\ack{
    We would like to thank Stefan Klus who contributed code for the double-well system, edits to old drafts, and many insightful conversations about continuous-time stochastic dynamical systems. 
    SLB acknowledges support from the Boeing Company, the National Science Foundation AI Institute in Dynamic Systems (grant number 2112085) and from the Army Research Office (ARO W911NF-19-1-0045).
}

\bibliographystyle{siam}
\bibliography{references}

\newpage
\appendix

\section{Reinforcement Learning Algorithm Definition}
\label{appendix:rl_algorithm_definitions}
The soft actor critic (Q) algorithm is provided in Alg.~\ref{alg:sacq}.   

\begin{algorithm}[H]
    \caption{Soft Actor-Critic (Q)}
    \label{alg:sacq}
    \begin{algorithmic}[1]
        \REQUIRE Initial parameters $\theta_1, \theta_2, \nu$
        \STATE Initialize target Q-network weights $\overline{\theta}_1 \gets \theta_1, \overline{\theta}_2 \gets \theta_2$
        \STATE Initialize replay pool $\mathcal{D}$ by interacting with the environment using the initial policy
        \FOR{$t \in$ iterations}
            \FOR{$s \in$ environment steps}
                \STATE $a \sim \pi_\nu(a | x)$ \COMMENT{Sample action}
                \STATE $(x', r(x, a)) \sim p(x' | x, a)$ \COMMENT{Sample transition}
                \STATE Store transition $(x, a, r(x, a), x')$ in $\mathcal{D}$
                \STATE $\mathcal{D} \gets \mathcal{D} \cup \{ (x, u, r(x, u), x') \}$
            \ENDFOR
            \FOR{$j \in$ gradient steps}
                \STATE $\theta_i \gets \theta_i - \lambda_Q \nabla_{\theta_i} J_Q(\theta_i)$ for $i \in \{1, 2\}$ \COMMENT{Update Q-function}
                \STATE $\nu \gets \nu - \lambda_\pi \nabla_\nu J_\pi(\nu)$ \COMMENT{Update policy weights}
                \STATE $\alpha \gets \alpha - \lambda \nabla_\alpha J(\alpha)$ \COMMENT{Adjust temperature}
                \STATE $\overline{\theta}_i \gets \tau \theta_i + (1 - \tau) \overline{\theta}_i$ for $i \in \{1, 2\}$ \COMMENT{Update target network}
            \ENDFOR
        \ENDFOR
        \ENSURE Optimized parameters $\theta_1, \theta_2, \nu$
    \end{algorithmic}
\end{algorithm}

\section{Episodic Returns: Linear System}
\label{appendix:episodic_returns_linear_system}
\begin{figure}[H]
    \centering
    \includegraphics[width=\textwidth]{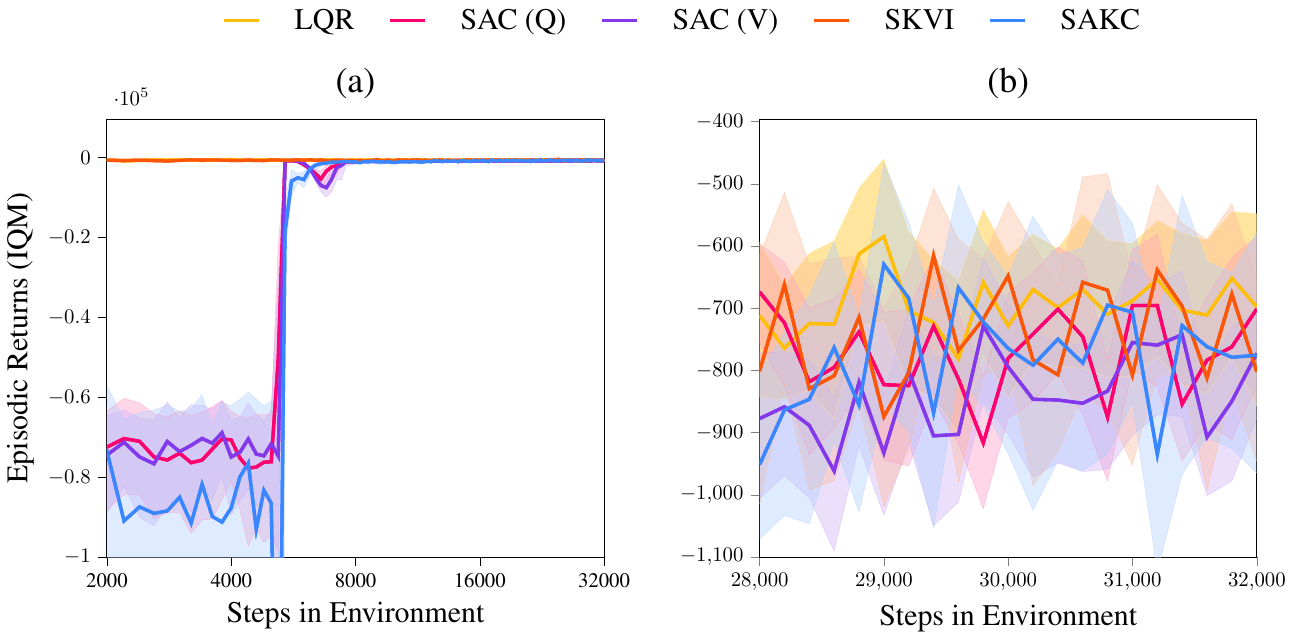}
    \caption{
        Inter-quartile means of the episodic returns on the linear system for \textbf{(a)} the entire control interval, and \textbf{(b)} a zoomed-in cutout of the last $4,000$ steps in the environment when all control algorithms have calibrated. As shown in the cutout, all control algorithms are within the confidence bands of each other and are as such statistically indistinguishable.
    }
    \label{fig:linear-system-iqm-returns}
\end{figure}

Performance of KARL on the linear system is shown in Fig.~\ref{fig:linear-system-iqm-returns}. In this example, all methods perform relatively well in steady state, although LQR and SKVI achieve higher episodic returns earlier. 

\vfill

\end{document}